\pdfoutput=1
\documentclass{article}
\usepackage{iclr2027_conference}
\usepackage{times}
\iclrfinalcopy
\usepackage{microtype}
\usepackage{graphicx}
\usepackage{subcaption}
\usepackage{booktabs}
\usepackage{pgfplotstable}
\usepackage{array}
\usepackage{longtable}
\usepackage{float}
\usepackage{placeins}

\usepackage{enumitem}
\usepackage{placeins}
\usepackage{amsmath}
\usepackage{dsfont}
\usepackage{amssymb}
\usepackage{pifont}
\usepackage{wrapfig}

\usepackage{xcolor}

\usepackage{snaptodo}

\snaptodoset{block rise=0.5em}
\snaptodoset{margin block/.style={font=\tiny}}
\renewcommand{\snaptodo}[2][1]{}

\usepackage{hyperref}
\usepackage{url}
\usepackage[utf8]{inputenc}
\usepackage[T1]{fontenc}
\usepackage{hyperref}
\usepackage{url}
\usepackage{booktabs}
\usepackage{amsfonts}
\usepackage{nicefrac}
\usepackage{microtype}
\usepackage{xcolor}
\colorlet{darkgreen}{green!45!black}
\hypersetup{
  colorlinks,
  citecolor=darkgreen}
\usepackage{hyphenat}

\usepackage{adjustbox}
\usepackage{amsfonts}
\usepackage{xcolor}

\usepackage{pifont}
\newcommand{\xmark}{\ding{55}}

\usepackage{graphicx}
\usepackage{enumitem}

\usepackage{amsmath, amssymb, mathtools, amsthm}
\usepackage{thmtools, thm-restate}

\theoremstyle{plain}
\declaretheorem{theorem}

\theoremstyle{definition}
\declaretheorem[sibling=theorem, numberwithin=section]{definition}

\theoremstyle{remark}

\newcommand{\defeq}{\overset{\text{\tiny def}}{=}}

\renewcommand{\d}{\rm d}
\renewcommand{\P}{\mathbb{P}}

\newcommand{\D}{\mathcal{D}}

\usepackage{bbold}
\usepackage{bigints}
\usepackage{stmaryrd}
\usepackage{booktabs,tabularx}
\usepackage{nicefrac, xfrac}
\usepackage{makecell}
\usepackage[nobiblatex]{xurl}
\newcommand\rurl[1]{%
  \href{http://#1}{\nolinkurl{#1}}%
}

\usepackage{stmaryrd}

\usepackage{tikz}
\usetikzlibrary{backgrounds}
\usetikzlibrary{tikzmark}
\usetikzlibrary{calc}
\usetikzlibrary{arrows,shapes,positioning,shadows,trees,mindmap}
\usetikzlibrary{arrows.meta}

\usepackage{tcolorbox}

\usepackage{mathtools}
\usepackage{amsthm}

\usepackage[capitalize,noabbrev]{cleveref}

\usepackage[textsize=tiny]{todonotes}

\makeatletter

\newcommand{\appendixtoc}{%
  \hspace*{.05\linewidth}%
  \begin{minipage}{.9\linewidth}
    \hrule\smallskip
    \@starttoc{apptoc}%
    \smallskip\hrule
  \end{minipage}
}

\newcommand{\enableappendixtoc}{%
  \renewcommand{\addcontentsline}[3]{%
    \def\apptoc@type{##1}%
    \def\apptoc@toc{toc}%
    \ifx\apptoc@type\apptoc@toc
      \addtocontents{apptoc}{%
        \protect\contentsline{##2}{##3}{\thepage}{\@currentHref}%
      }%
    \fi
  }%
}
\makeatother

\title{%
The Loss or the Architecture? What Matters in Survival Analysis with Competing Risks
}

\author{%
    Julie Alberge \\
  SODA Team\\
  Inria Saclay\\
  France \\
  \texttt{julie.alberge@inria.fr} \\
  \And
  Gaël Varoquaux \\
  SODA Team\\
  Inria Saclay\\
  France \\
  \And
  Judith Abécassis \\
  SODA Team\\
  Inria Saclay\\
  France \\
}

\begin{document}
\maketitle
\lhead{Preprint.}

\begin{abstract}
Often, \emph{eg} in healthcare and predictive maintenance, what matters is which of several competing causes will
strike first, and with what probability: survival analysis with competing risks. Corresponding deep learning methods are developed as bundles: a new architecture together with a new training loss, tested on one to five
datasets against a handful of baselines. Thus, a state-of-the-art result cannot be attributed to the
architecture, the loss, or the evaluation protocol. The coupling is not accidental: with one
incidence curve per cause, a network's head is tied to the losses expressible on it. We vary the two axes
independently, three losses, two of them strictly proper, on four backbones, across two modalities that
tease the axes apart: 20 tabular and 3 pathology-image datasets.
One loss wins in both: modelling cumulative incidence directly under
censoring weighting. It brings twice as much as benefits a good choice of backbone on competing-risks data, and no more than it on
single-event data: the loss is decisive for competing risks.
The backbone matters too, and must match the modality: tabular
networks lead on tables, pretrained vision encoders on images. The two choices compound rather than substitute.
Proper scoring rules reveal these dynamics; the concordance index (C-index), which is
not proper, does not separate models: neither axis moves it, not even tuning on the loss built to
improve it.
This is an avenue of progress: composed, the two give a model nobody has
tried, RealMLP or TabM trained with the cumulative-incidence loss, which takes the best Brier score in
our benchmark.
\end{abstract}

\section{Introduction: competing risks, and where the gains come from}

Time-to-event prediction arises when the quantity of interest is a delay: how long until a patient
relapses, a borrower defaults, a component fails. What makes it more than regression is
right-censoring: When follow-up ends, many subjects have simply not had the event yet, and all we know
is that their time exceeds it. Discarding them biases the model towards those who failed early;
counting them as event-free biases it the other way. Survival models instead use what a censored
subject does tell us, that the event came after a known time. They call for dedicated losses.

Competing risks adds a second twist: the event comes in several mutually exclusive kinds, and only the
first is ever seen, since a patient who dies of heart disease cannot later die of cancer. Treating the
other causes as further censoring and fitting one model per cause is tempting, but it answers the risk
in a world where those causes do not exist and so overstates the risk that matters. What predictions
need is the cumulative incidence function (CIF) of each cause: the probability that it strikes before
a given time and before any competing cause. Deciding whether to treat a patient or service a machine
builds on that absolute probability, more than finding who is most at risk.

Machine-learning research has produced many competing-risks methods, from semi-parametric estimators to random forests and neural networks (Sec.~\ref{sec:relatedwork}), and recent ones frequently introduce a new architecture and a new training loss at once \citep{nagpal_deep_2021, danks2022derivative, wang_survtrace_2022, jeanselme2023neuralfinegraymonotonicneural}. For example, \citet{lee_deephit_2018} pair a custom architecture with a likelihood-plus-ranking loss. Because the two axes are never varied separately, architecture and loss are confounded by construction, and a state-of-the-art result cannot be attributed to one or the other. The evidence base is also thin: Table~\ref{tab:benchmade} shows methods are introduced on one to five real competing-risks datasets and assessed with one to three metrics, not enough to separate models well. Often, the metric is concordance alone (C-index), which ranks individuals by risk and may miss differences in predicted probabilities, yet those absolute probabilities drive decisions in medicine \citep{van2019calibration}. The effect of proper scoring rules has been isolated in standard survival analysis on simple MLPs \citep{yanagisawa_proper_2023}, but competing risks has seen no such study.

\begin{wraptable}[14]{r}{0.52\textwidth}
    \footnotesize%
    \vspace*{-3.3ex}
    \caption{\textbf{Competing-risks methods are introduced on very thin evidence.} Datasets,
baselines and metrics used by the papers proposing methods, against this study.}\vspace*{-2.2ex}
    \begin{tabular}{lccc}
         & \hspace*{-2ex}\# Real Life CR\hspace*{-1ex}  & \# & \# CR  \\
         Publication & Datasets & baselines\hspace*{-1ex} & metrics\\
         \midrule
         \citealt{ishwaran_random_2014}\hspace*{-5ex} & 5 & 5 & 2\\
          \citealt{lee_deephit_2018}\hspace*{-5ex} & 1 & 3 & 1\\
          \citealt{wang_survtrace_2022}\hspace*{-5ex}  & 1 & 6 & 1 \\
         \citealt{jeanselme2023neuralfinegraymonotonicneural}\hspace*{-5ex}  & 3 & 5 & 3\\
         \citealt{alberge_survival_2025}\hspace*{-5ex}  & 1 & 7 & 3 \\
         \midrule
         \textbf{Ours} & \textbf{10} tabular & \hspace*{-3ex}\textbf{9 baselines +}\hspace*{-4ex} & \textbf{6} \\
         & \textbf{3} vision & \hspace*{-3ex}\textbf{12 loss}$\times$\textbf{arch.}\hspace*{-4ex} & \\
    \end{tabular}

    \label{tab:benchmade}
\end{wraptable}

We therefore build a benchmark that decouples the training loss from the backbone. Three losses are trained on shared neural backbones: a competing-risks likelihood over cause-specific event probabilities (HazardLikelihood), the hybrid ranking-plus-likelihood loss of DeepHit (HLRanking), and the cumulative-risk negative log-likelihood (CRNLL). The first and third are strictly proper scoring rules under their respective assumptions, meaning that only the true predictive distribution minimizes them; the ranking term makes HLRanking improper. We evaluate them on 10 competing-risks and 10 single-event tabular datasets (including SEER \citep{seer_dataset}, MIMIC \citep{PhysioNet-mimiciii-1.4, PhysioNet-mimiciv-3.1}, eICU \citep{pollard_eicu_2018}, and METABRIC \citep{metabric}) with a randomized hyperparameter search for each training scenario, and report probabilistic accuracy, discrimination, and calibration.

We then adapted the whole comparison on pathology images, with pretrained vision backbones \citep{resnet, densenet, krizhevsky2012imagenet}. This change of modality introduces new variation. Indeed, for tables gains from the backbones are small, so the architecture axis is narrow, while the loss reshapes what the whole network is fit to. The second modality, vision, tells the two axes apart cleanly, because it moves one of them a long way while leaving the other alone: we use the same three losses, whereas the backbone explores architecture dedicated to vision (pretrained encoder) as well as a general one, an MLP. The ordering of the losses survives the move, while the backbone that wins does not: pretrained encoders lead on images, tabular networks on tables.

\paragraph{Contributions}
\begin{description}[nosep, leftmargin=*]
    \item[A controlled loss $\times$ architecture matrix:] Each of three losses trained on each of four backbones, with the same search space and the same 150-iteration budget in every cell, so the two axes can be read off independently; and nine methods with their own fixed losses serve as baselines.
    \item[An evidence base large enough to separate two effects:] 10 competing-risks and 10 single-event tabular datasets plus three built from pathology images (an order of magnitude more than the studies of Table~\ref{tab:benchmade}) scored on metrics spanning accuracy, discrimination and calibration.
    \item[The loss decides in competing risks, and properness does not suffice:] Modeling cumulative incidence directly under censoring weighting improves strongly on both hazard-based losses on tabular data, and still leads on every frozen vision encoder. It brings twice as much benefits as a good backbone in competing-risks settings, while in single-event settings the two contribute equally to the performance. Two strictly proper losses, CRNLL and HazardLikelihood, nonetheless sit far apart, so properness is needed to trust a metric but not sufficient to make a loss competitive. The one improper ingredient we test, DeepHit's ranking term, is down-weighted by tuning.
    \item[The backbone must match the modality:] RealMLP and TabM lead on tables, pretrained encoders on images, and the MLP under-performs on images whatever the loss.
    \item[Ranking metrics cannot see any of this:] Nothing measurably moves the C-index, not even tuning on the loss built to optimize it. A study reporting concordance alone, which Table~\ref{tab:benchmade} shows is common practice, would have concluded that none of these choices matters.
    \item[A stronger model, obtained by composition:] Pairing a recent tabular architecture with the cumulative-incidence loss, which no prior work has done, beats every published neural competing-risks method on the Brier score,
and meet or exceed results of SurvivalBoost, the tree-based method in our benchmark that builds on the same loss.
\end{description}

\paragraph{Scope} We compare losses and backbones for competing-risks prediction rather than proposing new architecture. Some losses impose a parametric form on the CIF \citep[eg Deep Survival Machines, ][]{nagpal_deep_2021}; they cannot be added onto an arbitrary backbone and so enter as fixed-loss baselines rather than as cells of the matrix. Time-varying covariates and recurrent events are out of scope.

\section{Setting: competing risks, losses, and how to score them}
In the following, we note random variables in capitalized letters, and observations are in lowercase.

\subsection{Competing risks: one incidence curve per cause}
We follow the standard formulation of the competing-risks problem \citep{pintilie2006competing}, with $K > 1$ types of event. Let $T^\star \in \mathbb{R}_+$ be a continuous random variable representing the time to the first event, and $\Delta^\star \in \{1, \dots, K\}$ be a categorical variable denoting the specific event. We also face right-censoring, denoted~$C$. The final time observation is $T = \min( T^\star,C)$. For a given sample, the observed triplet $(\mathbf{x}, (t, \delta))$, where $\mathbf{x} \in \mathbb{R}^d$ denotes the $d$ covariates, $t$ the time observed and $\delta$ indicates the event type (with $\delta = 0$ representing censoring). The primary quantity of interest is the Cumulative Incidence Function (CIF) for each cause $k$, defined as $F_k(t, \mathbf{x}) = \mathbb{P}(T^\star \leq t, \Delta^\star = k \mid \mathbf{X}=\mathbf{x})$. It can be modelled directly \citep{aalen_empirical_1978, lee_deephit_2018, alberge_survival_2025}, or through the cause-specific hazard or density, its derivative \citep{fine_proportional_1999, kvamme_continuous_2019, wang_survtrace_2022, jeanselme2023neuralfinegraymonotonicneural}. Appendix~\ref{sec:computations} gives the computations.

\subsection{Three losses that any backbone can carry, two of them proper}\label{sec:crlosses}
We consider losses imposing no strong parametric prior on the CIF, so they can be moved across backbones. Three qualify: two strictly proper scoring rules and the composite loss of \citet{lee_deephit_2018}.

\paragraph{HazardLikelihood Loss} The most widely adopted loss is the competing-risks negative log-likelihood, a strictly proper scoring rule \citep{guan2021properscoringrulevalidation} used in several works \citep{fine_proportional_1999, jeanselme2024review}. It is also what the SurvTRACE repository implements~\citep{wang_survtrace_2022}, although the original manuscript discusses alternatives.

\begin{definition}[Hazards Likelihood]\label{def:losslikelihood}
The log-likelihood in competing risks setting can be defined as:
\begin{equation}
\begin{aligned}
\mathcal{L}_{HL}\triangleq- & \sum_{i=1}^N\Biggl [ \mathbb{1}_{\delta_i \neq 0} \cdot \log \left(\hat{f}_k(t_i|\mathbf{x}_i)\right)\,
 + \,
\mathbb{1}_{\delta_i =0} \cdot \log \left(1-\sum_{k=1}^K \hat{F}_k\left(t_i \mid \mathbf{x}_{i}\right)\right)\Biggr ],
\end{aligned}
\end{equation}

where $\delta_i \in \{0, 1, \dots, K\}$ is the event indicator (with $0$ denoting right-censoring), $\hat{f}_{\delta_i}$ is the predicted cause-specific density for the realized event, and $\hat{F}_k$ is the predicted CIF for cause $k$. As shown by \citet{guan2021properscoringrulevalidation}, this loss is a proper scoring rule.
\end{definition}

\paragraph{HazardLikelihoodRanking Loss}
The loss used to train DeepHit \citep{lee_deephit_2018}, which we call HazardLikelihoodRanking (HLRanking), adds a ranking term to that likelihood. Following its pycox implementation \citep{kvamme2019pycox}, $\mathcal{L}_{HLR} = \alpha\,\mathcal{L}_{HL} + (1-\alpha)\,\mathcal{L}_{R}$ with a hyperparameter $\alpha\in[0,1]$, where $\mathcal{L}_{HL}$ is the negative log-likelihood of Def.~\ref{def:losslikelihood} and $\mathcal{L}_{R}$ is a ranking loss (Def.~\ref{def:lossranking}).

\begin{definition}[Ranking Loss]\label{def:lossranking}
Let $A_{k, i, j} \triangleq \mathbb{1}\left(\delta_i=k, t_i<t_j\right)$ be an indicator that patient $i$ experiences event $k$ before patient $j$ experiences any event or is censored. It is defined as:
\begin{equation}
\mathcal{L}_R \triangleq \sum_{k=1}^K \sum_{i \neq j} A_{k, i, j} \cdot \eta\left(\hat{F}_k\left(t_i \mid \mathbf{x}_{i}\right), \hat{F}_k\left(t_i \mid \mathbf{x}_j\right)\right)
\end{equation}
where $\eta(x, y)$ is a convex penalty function (e.g., exponential) that penalizes incorrect risk orderings.
\end{definition}

The ranking penalty renders HLRanking improper. Note however that tuning on IBS selects $\alpha$ values which downplay the ranking penalty and pulls HLRanking towards $\mathcal{L}_{HL}$ (Appendix Tab.~\ref{tab:alphaselected}).

\paragraph{CompetingRisksNegativeLogLikelihood Loss}
The following loss is designed to directly optimize the CIFs, and can be seen as the negative log-likelihood, re-weighted by the censoring distribution using Inverse Probability of Censoring Weighting (IPCW), which up-weights the subjects still under observation so that they stand in for those already lost to follow-up. This is a strictly proper scoring rule \citep{alberge_survival_2025} under the assumption that the censoring weights are correctly specified.

\begin{definition}[CompetingRisksNegativeLog-Likelihood (CRNLL)]
    $\quad\forall \tau, (\mathbf{x}, t, \delta) \sim \mathcal{D}, $%
    \begin{multline}
    \mathcal{L}_{CRNLL, \tau}\left(\left(\hat{F}_1(\tau \mid \mathbf{x}), \dots, \hat{F}_K(\tau \mid \mathbf{x}), \hat{S}(\tau \mid \mathbf{x})\right), (t, \delta)\right) \triangleq \\
    \frac{1}{n} \sum_{i=1}^n \left[ \sum_{k=1}^{K}
    \dfrac{
        \mathbb{1}_{t_i \leq \tau, \delta_i = k} \log\left(\hat{F}_k(\tau \mid \mathbf{x}_i)\right)
        }
        {G^\star(t_i \mid \mathbf{x}_i)
        } \right.
        \left. +
        \dfrac{
        \mathbb{1}_{t_i > \tau}\log\left(\hat{S}(\tau \mid \mathbf{x}_i)\right)
        }
        {G^\star(\tau \mid \mathbf{x}_i)
        } \right]
    \label{eqn:full_loss}
\end{multline}
where $G^\star$ is the complement of the CDF of the censoring distribution, i.e. the probability of not being censored for a given sample until the time $t_i$ or $\tau$.
\end{definition}

\subsection{Metrics: probabilistic accuracy, discrimination, calibration}\label{sec:crmetrics}
Competing-risks predictions must be right about both which event occurs first and when, under
right-censoring. Following \citet{qi_toward_2025}, we study six evaluation metrics along three axes, capturing
different aspects of predictive performance; Appendix~\ref{sec:metrics} defines each one of them.

\emph{Probabilistic accuracy:} the Integrated Brier Score \citep{graf_assessment_1999}, the
time-integrated squared error between the predicted cumulative incidence and the observed outcome, and
the Integrated Negative Log-Likelihood, its log-loss analogue; both are strictly proper scoring rules
under the assumptions discussed next.
\emph{Discrimination:} the competing-risks C-index \citep{harrell1996multivariable, antolini_timedependent_2005, uno_cstatistics_2011,
wolbers_prognostic_2009}, measuring whether patients are ranked in the right order, which we evaluate at several time horizons since the ranking may change in time; and the
cumulative/dynamic time-dependent AUC \citep{uno_evaluating_2007, blanche_estimating_2013}.
\emph{Calibration:} the competing-risks extensions of D-calibration \citep{haider2020effective} and of
Aalen--Johansen calibration \citep{haider2020effective, dagostino_evaluation_2003} introduced by
\citet{alberge2026calibrationsurvivalmodelscompeting}, which detect systematic over- or
under-estimation of a cause's cumulative incidence.

\paragraph{Prioritizing the IBS/INLL}
We focus on the Integrated Brier Score rather than the C-index, for two reasons. First, the IBS is a strictly proper scoring rule \citep{gneiting_strictly_2007, rindt_survival_2022, yanagisawa_proper_2023} under two assumptions: censoring is independent of covariates and that the censoring distribution is correctly estimated by the inverse probability of Censoring Weighting (IPCW). It is then minimized if and only if the predicted distribution matches the true underlying distribution.
The C-index, in contrast, has drawn criticism \citep{rindt_survival_2022, lillelund2025stopchasingcindexevaluate}: being a pure ranking metric it is blind to the magnitude of the predicted risks, and \citet{rindt_survival_2022} show it can score an incorrect distribution above the oracle, making it improper.

\section{Related work: many methods, few controlled comparisons}\label{sec:relatedwork}

\paragraph{Benchmarks} Standard survival analysis has seen several benchmarks in the past few years \citep{yanagisawa_proper_2023, jeanselme2024review, wang2024survivalmodelingusingdeep, mesinovic2025survbenchstandardisedpreprocessingpipeline, noroozizadeh2026survhtebenchbenchmarkheterogeneoustreatment}, supported by libraries packaging datasets \citep{mesinovic2025survbenchstandardisedpreprocessingpipeline, kvamme2019pycox, nagpal2022auton, drysdale2022} or metrics \citep{qi_survivaleval_2024, sksurv}, and reviewed by \citet{wiegrebe2024deep}. Closest to us, \citet{burk2026neutral} neutrally compare 19 single-event models over 34 datasets, tuning on both a ranking measure and a scoring rule as we do, and \citet{kopper2024training} vary the scoring rule used for \emph{training} single-event models, reporting only minor differences between rules. Competing risks has been studied far less, for want of easily obtained datasets: SurvTRACE \citep{wang_survtrace_2022}, for instance, is benchmarked on a single real dataset. Reviews \citep{djangang2025comparative} and benchmarks \citep{van_geloven_validation_2022, sierra2026reproducible} have begun to fill the gap, though on few datasets and without varying the loss.

\paragraph{Modern Methods} Methods range from the Aalen--Johansen estimator \citep{aalen_empirical_1978} to support vector machines \citep{van_belle_support_2011}, Gaussian processes \citep{guassianprocesses}, trees \citep{ishwaran_random_2014, bellot_tree-based_2018}, neural networks \citep{lee_deephit_2018, nagpal_deep_2021, katzman2018deepsurv} and transformers \citep{wang_survtrace_2022, ries2025tracertransformerbasedcompetingrisk}. Deep competing-risks methods each introduce a network together with a likelihood-type loss of their own \citep{lee_deephit_2018, nagpal_deep_2021, danks2022derivative, jeanselme2023neuralfinegraymonotonicneural}; none compares losses across backbones, which is the comparison we run.

\section{Building blocks: one adaptation that fits every backbone}

\begin{figure}[t]
    \centering
    \includegraphics[width=\linewidth]{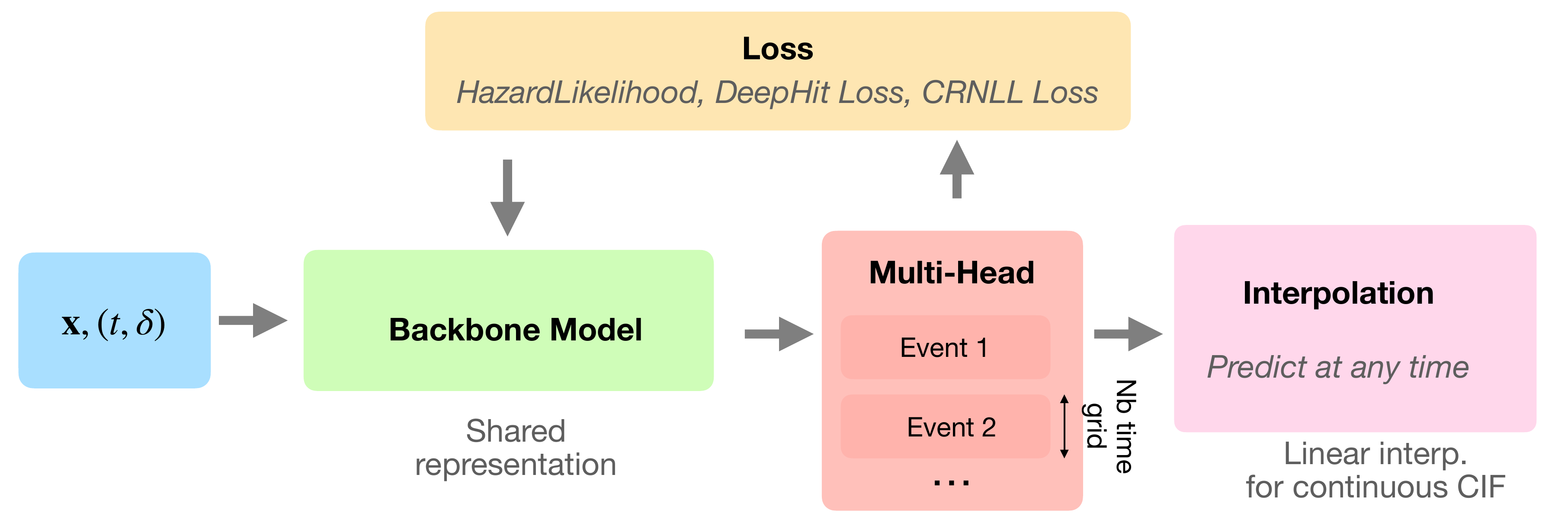}
    \caption{\textbf{Any backbone becomes a competing-risks model with two changes.} One prediction
    head per cause, plus a survival head for losses that model the cumulative incidence directly,
    and a discretized time grid at the output. Everything upstream is left as the original
    architecture, which is what lets us vary loss and backbone independently.}
    \label{fig:backbonetocr}
\end{figure}

We adapt a neural network backbone to the CR setting with the minimum changes, shown in Fig.~\ref{fig:backbonetocr}.

\paragraph{Multi-Head Architecture:} Competing risks need one incidence curve per cause.
The backbone becomes a shared representation learner branching into one head per curve.
Hazard-based losses use $K$ heads, and the overall survival function follows by integrating the hazards over time, as in DeepHit and its pycox implementation \citep{kvamme2019pycox}. For the CRNLL loss, which models the CIF directly, we use $K+1$ heads followed by a softmax activation to ensure valid, normalized probability distributions across all events for each sample. Because this additional survival head is a parameterization choice rather than a property of the CRNLL loss itself, we run an ablation isolating the two: CRNLL trained with the extra $(K+1)$-th head against CRNLL trained with only $K$ heads (renormalized without a dedicated survival output), on all four backbones. Removing the head shifts neither IBS nor calibration systematically on any backbone, so what we report is an effect of the loss, not of the output parameterization (Appendix~\ref{sec:survivalheadablation}).

\paragraph{Time-Grid Discretizations and Continuous Inference:} Training uses a discrete time grid built from observed event times, excluding censored ones; its resolution is a tuned hyperparameter (20 to 60 points, Appendix Tab.~\ref{tab:hprealmlp}). At inference we interpolate linearly between grid points.

\paragraph{Specificity for each loss} For CRNLL, we use the marginal \citet{kaplan_nonparametric_1958} estimator for the training-time IPCW, rather than a covariate-conditional Cox estimator, which a sensitivity analysis (Sec.~\ref{sec:ipcwsensitivity}) found fragile: it converges on only a dataset-dependent subset of our benchmark, and whenever its IBS differs from the Kaplan-Meier one, it is always worse, never better.

\paragraph{Backbone model}
We train every loss on four backbones, each adapted as in Fig.~\ref{fig:backbonetocr}: a plain 3-layer MLP, which is DeepHit's own architecture \citep{lee_deephit_2018} and which we call MLP throughout; RealMLP \citep{holzmuller2024better}, an MLP tuned for tabular data; TabM \citep{gorishniy2025tabm}, which imitates an ensemble of MLPs; and the SurvTRACE transformer \citep{wang_survtrace_2022}. RealMLP and TabM are the two best non-foundation models on TabArena \citep{tabarena2024}.

\paragraph{Vision backbones} The adaptation of Fig.~\ref{fig:backbonetocr} is not specific to tabular
data, so we apply it unchanged to image inputs. We use three standard pre-trained architectures (AlexNet \citep{krizhevsky2012imagenet},
MobileNetV2 \citep{mobilenet} and ResNet50 \citep{resnet}) from \texttt{torchvision}
\citep{torchvision2016} with their default weights. To isolate the effect of the competing-risks
adaptation from that of representation learning, we freeze the backbone parameters and use the networks
as fixed feature extractors, optimizing only the heads. We also carry the tabular MLP over unchanged,
which tests whether a backbone has to match the data.

\section{Experiments: 23 datasets, equal tuning budget in every cell}
\subsection{Datasets: 10 competing-risks, 10 single-event, 3 image}

\paragraph{Tabular data}
We benchmark on 10 competing-risks tabular datasets with two or three competing events and, since a
competing-risks model handles standard survival analysis as the case $K=1$ while single-event datasets
are far more plentiful, on 10 single-event datasets as well, matching the two counts. They serve as a
contrast as much as a supplement: the same losses and backbones on a problem without competing causes
isolate what the competing-risks structure itself contributes (Appendix~\ref{sec:effectsizes}). All have more than 1,000
samples. This collection is the larger of the two because most competing-risks data is tabular, clinical
registries and cohort studies above all. Table~\ref{tab:datasetsconsidered} summarises them; Appendix~\ref{sec:datasets} gives for
each one how it was obtained and preprocessed together with its full characteristics, and
Appendix~\ref{sec:studymimic} states the research question and censoring definition adopted for
MIMIC-III/IV \citep{PhysioNet-mimiciv-3.1, PhysioNet-mimiciii-1.4} and eICU
\citep{pollard_eicu_2018}.

\begin{table}[b!]
    \caption{\textbf{Our 20 tabular survival datasets, spanning three orders of magnitude in sample size and censoring
    rates from 19\% to 95\%.} The three image datasets are described in Sec.~\ref{sec:visiondata}. Appendix~\ref{sec:datasets}
    gives the full characteristics and the access procedure for each one.}
    \label{tab:datasetsconsidered}
\begin{minipage}{.47\linewidth}
    \centering
    \small
    \begin{tabular}{lrc}
    \multicolumn{3}{c}{Competing Risks}  \\
    \toprule
    Dataset & Samples & Cens.  \\
    \midrule
    Framingham      & 4,240      & 58\% \\
    Metabric (CR)   & 3,237      & 33\% \\
    MIMIC IV        & 65,366     & 58\% \\
    MIMIC III       & 38,511     & 54\% \\
    SEER Breast     & 476,746    & 63\% \\
    SEER Digestion  & 887,552    & 19\% \\
    PBC             & 1,945      & 55\% \\
    Loan            & 1,652,304  & 41\% \\
    eICU            & 106,196    & 95\% \\
    TCGA            & 10,540     & 70\% \\
    \bottomrule
    \end{tabular}
\end{minipage}%
\hfill%
\begin{minipage}{.47\linewidth}
    \centering
    \small
    \begin{tabular}{lrc}
     \multicolumn{3}{c}{Survival Analysis} \\
    \toprule
    Dataset & Samples & Cens. \\
    \midrule
     Flchain      & 6,524      & 70\% \\
     Hdfail       & 52,422     & 95\% \\
     GBSG         & 2,232      & 43\% \\
     Metabric     & 1,904      & 42\% \\
     MSK 2021     & 25,659     & 60\% \\
     MSK 2026     & 40,755     & 57\% \\
     NWTCO        & 4,028      & 85\% \\
     Smarto       & 3,873      & 88\% \\
     Support      & 8,873      & 32\% \\
     ENS Challenge& 3,173      & 50\% \\
    \bottomrule
    \end{tabular}
\end{minipage}
\end{table}

\paragraph{Image data}\label{sec:visiondata}
Beyond the tabular collection, we assembled three competing-risks datasets from TCGA whole-slide
pathology images \citep{lu2021data, kandoth2013mutational}, preprocessed following
\citet{liu10385148}, giving 635 images. Combining these with the clinical records from
cbioportal\footnote{\url{https://www.cbioportal.org}}, we defined two competing events on each of
them: death from cancer and from other causes.

\subsection{Tabular baselines: nine methods with their own losses}
The four backbones described above (MLP, RealMLP, TabM, SurvTRACE) are each trained under all three losses of Sec.~\ref{sec:crlosses}, forming our controlled loss $\times$ architecture matrix; the hyperparameter grids are in Appendix Tab.~\ref{tab:hprealmlp}. We also include baseline models with different learning paradigms, each trained with its own original, fixed loss rather than as part of that matrix.
\begin{description}[leftmargin=2em]
    \item[Aalen--Johansen] \citep{aalen_empirical_1978}: the covariate-free, population-level estimator of the CIFs; the reference of all our comparisons.
    \item[Fine and Gray] \citep{fine_proportional_1999}: a linear model of the subdistribution hazard, under proportional hazards.
    \item[Random Survival Forests] \citep{ishwaran_random_2014}: random forests adapted to competing risks.
    \item[SurvivalBoost] \citep{alberge_survival_2025}: gradient-boosted trees trained with the CRNLL loss and a feedback loop on the censoring model.
    \item[DeepHit] \citep{lee_deephit_2018}: a neural network trained with the HLRanking loss (Def.~\ref{def:lossranking}).
    \item[Deep Survival Machines (DSM)] \citep{nagpal_deep_2021}: a mixture of parametric distributions (e.g.\ Weibull) for the CIFs, fit by maximum likelihood.
    \item[DeSurv] \citep{danks2022derivative}: each cause's CDF as the integral of a neural network, valid by construction, fit by maximum likelihood.
    \item[Neural Fine-Gray] \citep{jeanselme2023neuralfinegraymonotonicneural}: Fine and Gray with monotonic neural networks, relaxing proportional hazards while keeping valid CIFs.
    \item[SurvTRACE] \citep{wang_survtrace_2022}: a transformer whose published code implements only the `naive' loss, i.e.\ HazardLikelihood (Def.~\ref{def:losslikelihood}).
\end{description}

\begin{figure}[t]
    \centering
    \begin{subfigure}{\linewidth}
        \centering
        \includegraphics[width=\linewidth]{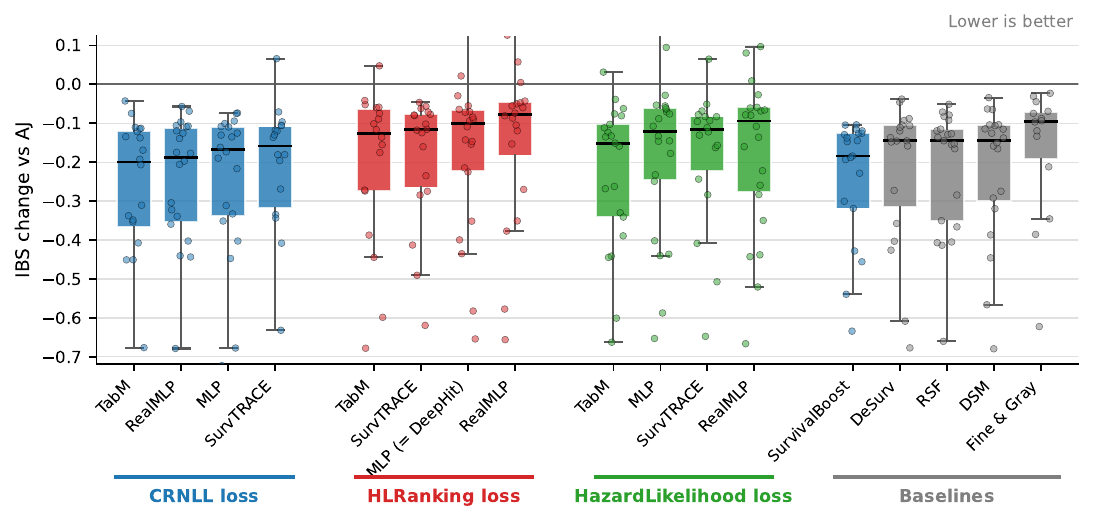}%
	\vspace*{-1ex}
        \caption{\textbf{20 tabular datasets}: IBS gain over Aalen--Johansen, higher is better. Four backbones
        per loss, plus nine baselines carrying their own fixed losses.  One box per model, one
        point per dataset, box spanning the interquartile range. SurvivalBoost is a non-differentiable baseline that uses the CNRLL loss.}
        \label{fig:ibsgain}
    \end{subfigure}

    \begin{subfigure}{\linewidth}
	\hspace*{.01\linewidth}%
        \includegraphics[width=0.71\linewidth]{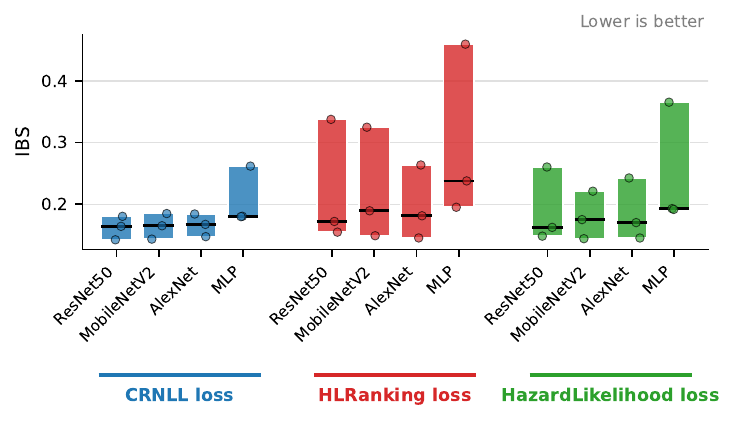}%
	\vspace*{-1ex}
        \caption{\textbf{3 TCGA pathology-image datasets}: raw IBS, lower is better. Three frozen pretrained encoders and the tabular MLP, no
        external baselines.}
        \label{fig:visiongain}
    \end{subfigure}
    \caption{\textbf{A competitive competing-risks model needs a backbone matched to the modality
    \emph{and} a loss that models cumulative incidence under censoring; neither substitutes for
    the other.} Both panels are grouped and coloured by loss, with the backbone named beneath each
    box. In both modalities CRNLL comes first within every backbone. What changes is which backbone
    wins: on tables the backbones built for tables, RealMLP and TabM, lead the plain MLP, while on
    images the pretrained encoders lead it by a wide margin and the MLP is also the least stable. The
    two panels have different backbones, baselines and vertical axes, and are not to be compared cell by
    cell.}
    \label{fig:central}
\end{figure}

\subsection{Evaluation strategy: two hyperparameter searches per cell}
Each dataset is randomly split into three folds (train + validation sets: $70\%$, and test set: $30\%$) and each split is capped at 50,000 samples. Each method is trained using \texttt{RandomizedSearchCV} from scikit-learn \citep{scikit-learn} with 150 iterations. Selecting on validation IBS alone would leave models undertuned for the C-index, which is one measure of our benchmark. We therefore run the search \emph{twice} in every cell (pair of loss \& architecture), once tuning for IBS and once tuning for C-index, and evaluate the two selections separately. This matters most for HLRanking, whose weight $\alpha$ between likelihood and ranking and ranking temperature $\sigma$ are searched with the other hyperparameters (Tab.~\ref{tab:hprealmlp}): the C-index search is free to lean on the ranking term. IBS is reported as a gain over the Aalen--Johansen estimator, $1 - \mathrm{IBS}/\mathrm{IBS}_{\mathrm{AJ}}$, putting datasets on a common scale.
After choosing the hyperparameters, we evaluate the models on 5 train/test splits.

\section{Results: the loss and the backbone each matter}\label{sec:results}

\paragraph{Trends across the two modalities}
Figure~\ref{fig:central} gives results for tabular (top panel) and image (bottom panel) data, for IBS.
The two panels have different backbones and different baselines. But they show common trends.
They agree on the loss: in both modalities, and within every backbone, modeling the
cumulative incidence directly under censoring weighting comes first. Where they differ is in which
backbone wins, and this difference is itself informative. On tables the backbones designed for tables
lead, RealMLP and TabM ahead of the plain MLP; on images the pretrained encoders lead and that same MLP
is last, and least stable.
Neither choice stands in for the other: a good loss on a mismatched backbone, and a matched backbone
with a weaker loss, are both beaten by the pair.

\paragraph{Tabular data: the loss is decisive with competing risks}
We compare the effect of loss versus backbone: the spread across losses
holding the backbone fixed, and the converse (Appendix~\ref{sec:effectsizes}). Here we observe different trends in competing-risks versus single-event survival.
On the 10 competing-risks datasets the loss effect is $0.053$ against $0.027$ for the
backbone, so the loss moves the Brier score about twice as much. On the 10 single-event datasets the
order reverses, $0.072$ against $0.086$, the two axes being comparable and the backbone marginally
ahead. Pooled across the two families of datasets, the effects average out
(Tab.~\ref{tab:effectsizes}). The loss therefore matters most in competing risks settings; modeling jointly the several mutually-exclusive causes to censoring makes getting it right decisive. Table~\ref{tab:lossbackbonemedians} shows the same
contrast cell by cell: on competing risks CRNLL leads the nearest hazard-based loss by a wide margin on
every backbone, whereas on single-event data the three losses fall closer together, in line with \citet{kopper2024training}. On competing risks, swapping DeepHit's loss for CRNLL on DeepHit's own MLP more than doubles its median IBS gain, from $0.12$ to $0.26$.

Figure~\ref{fig:ibsgain} shows the distribution, across the 20 datasets, of the IBS gain over the Aalen--Johansen estimator, the covariate-free empirical incidence curve; the competing-risks datasets alone (Fig.~\ref{fig:ibsgaincronly}) and the most heavily censored ones (Fig.~\ref{fig:ibsgainhighcens}) put the losses in the same order, with wider margins. By IBS, CRNLL comes first among the losses: it beats each hazard-based loss on all 20 datasets (median over backbones; sign test $p<10^{-6}$) and in about 90\% of (dataset, backbone) cells, while the two hazard-based losses tie. That tie is not a coincidence: tuning down-weights the ranking term (Sec.~\ref{sec:crlosses}), so HLRanking resembles HazardLikelihood.  Properness does not account for the ordering either, since CRNLL and HazardLikelihood are both strictly proper yet sit far apart: what sets CRNLL apart is that it models the cumulative incidence directly and reweights for censoring, rather than using cause-specific hazards. Properness matters for a metric to give a sharp picture (Appendix~\ref{sec:allmetrics}), but does not suffice to make a loss competitive.
Once the loss is fixed, architecture still matters, as the gap between RealMLP and MLP within the CRNLL group shows. On tabular data the loss is crucial, though it needs a \emph{tabular} backbone.

\paragraph{Pathology images: the backbone decides, the loss still matters}
Here the twelve cells are the three losses on three pretrained vision encoders and on the same MLP
(Fig.~\ref{fig:visiongain}). There are no external baselines as competing risks has been less studied for images. The backbone now separates the models far more than on tables: every
pretrained encoder beats the MLP under every loss, and the MLP's spread across datasets is several
times wider, so a tabular-style backbone on images is not merely worse but unreliable. The ordering of
the losses is nevertheless the one we measured on tables, CRNLL first within each backbone, then
HazardLikelihood, then HLRanking, including for the MLP. It holds although the encoders are frozen: acting on the heads
alone, the loss still orders the models.

\paragraph{C-index}
Neither loss nor architecture has visible effect on the C-index: every method performs alike (Fig.~\ref{fig:cindex25gain}). Even HLRanking, built for this metric, gains only marginally at the 25th percentile and nothing at the 50th or 75th. All quantiles (Fig.~\ref{fig:cindexquantiles}) and the integrated AUC (Fig.~\ref{fig:aucgain}) agree.

\paragraph{Does tuning HLRanking on its own target metric change the picture?}
Appendix Tab.~\ref{tab:hlrankingselection} compares HLRanking's own IBS and C-index (50th quantile) under the two selection criteria, paired at the (dataset, backbone) level. Pooled over the four backbones, selecting on C-index improves the C-index by a median of $+0.0018$ and degrades IBS by $+0.0117$: the gain is small and it is paid for dearly. This is the same finding seen from the other side. Left free, the search down-weights the ranking term; forced to favour concordance, it up-weights it (median $1-\alpha$ from $0.41$ to $0.71$, Tab.~\ref{tab:alphaselected}) and buys a negligible amount of C-index at a real cost in Brier score.

\paragraph{Calibration gives little signal}
Neither calibration criterion separates the losses usefully, and the two disagree: CRNLL is best
calibrated on AJ-calibration (Fig.~\ref{fig:aj_calib_raw}) and trails the hazard-based losses on
D-calibration (Fig.~\ref{fig:dcalib}). The likely reason is that there are multiple ways of defining calibration for competing risks, all imperfectly capturing errors to true probability, and all difficult to estimate. We report both AJ and D calibration for completeness but draw no conclusion from them.

\section{Conclusion: backbone for the modality, loss for the censoring}\label{sec:discussion}

\paragraph{Limitations} \emph{Not pure competing-risks.} Half of the 20 tabular datasets are
single-event ($K=1$). The ordering of the losses is the same on either family
(Fig.~\ref{fig:ibsgaincronly}), but the effect sizes are not, and we report the two apart wherever that
matters (Appendix~\ref{sec:effectsizes}).
\emph{Sample size.} Splits are capped at 50{,}000 training samples, so the largest datasets never
exercise the large-data regime.

\paragraph{Combining loss and backbone for the win} We presented a large-scale benchmark for survival
analysis with competing risks, spanning loan default, long-term medical follow-up and ICU outcomes. On
tabular data, the cumulative-incidence loss (eq.\,\ref{eqn:full_loss}) is the best of the three losses on every
backbone and, pooling backbones, on all 20 datasets; it is worth about twice the backbone where competing causes are present, against a
comparable weight on single-event data. Proper losses help single-event models \citep{yanagisawa_proper_2023}, but properness alone does not explain this, since the hazard
likelihood is proper too: what matters is modeling the cumulative incidence directly under censoring
weighting. The same loss still leads on pathology images, so its advantage is not an artifact of the
tabular regime. The backbone, in turn, has to match the modality: the MLP that is competitive on tables
is the worst and least stable model on images. A practitioner should thus pick the backbone for the
data and the loss for the censoring. Adapting the loss to censoring may seem obvious, yet recent deep
competing-risks methods have invested in dedicated architectures and output parameterizations \citep{lee_deephit_2018, nagpal_deep_2021, danks2022derivative, wang_survtrace_2022, jeanselme2023neuralfinegraymonotonicneural, ries2025tracertransformerbasedcompetingrisk}, perhaps because several incidence curves seem
to call for a dedicated network, and none builds on recent tabular architectures. Our results argue for undoing this coupling: one head per cause on a
generic backbone suffices, and the gains lie in the loss. Pairing RealMLP or TabM, built for tabular prediction, with the
cumulative-incidence loss beats every published neural competing-risks model on the Brier score. Advances in backbones will keep
arriving from outside this field; they transfer only once paired with a loss that suits the task, and
establishing whether they do calls for rigorous, many-dataset benchmarks such as this one. We will release
its code, dataset loaders and search spaces, so that a new backbone or loss can be scored under the
same protocol.

\paragraph{Only strictly proper scoring rules cleanly separated the models}
Of our six evaluation metrics (Appendix~\ref{sec:allmetrics}), only the two strictly proper scoring
rules gave clear signals: the Brier score and the log-likelihood order the models consistently and
alike. The ranking metrics, C-index and AUC, move for nothing, not even for the loss built for
concordance, tuned on it; the calibration criteria disagree. Theory
agrees: the C-index is not proper \citep{rindt_survival_2022, lillelund2025stopchasingcindexevaluate}
and the AUC not strictly so, so gaining on either need not mean a better distribution. We recommend
comparing models on a strictly proper scoring rule, keeping the rest as diagnostics. Applications that
need only a ranking lose nothing by it: our best loss is competitive on the C-index anyway.

\newpage
\subsection*{Acknowledgements}

JA acknowledges HPC resources from GENCI–IDRIS (Grant AD011017641).
JA, JA, and GV acknowledge funding from the European Union’s Horizon Europe research and innovation program under grant agreement No 1010954433 (Intercept-T2D). GV acknowledges support from ANR via grant TaFoMo (ANR-25-CE23-1822). This work is partly supported by Hi! PARIS and ANR/France 2030 program (ANR-23-IACL-0005).

\subsection*{Reproducibility statement}
The full benchmark codebase -- loss implementations, backbone architectures, hyperparameter search spaces, evaluation logic, and dataset preprocessing/splitting will be released on GitHub after publication. Section~\ref{sec:datasets} and Appendix~\ref{sec:availability} detail how each dataset was obtained and preprocessed, including access requirements for credentialed sources. Appendix~\ref{sec:computations} gives the full derivations for every loss and calibration metric used in the main text, and Tab.~\ref{tab:hprealmlp} reports the hyperparameter search space and compute budget for every model.

\FloatBarrier
\bibliography{benchmark}

\begin{thebibliography}{73}
\providecommand{\natexlab}[1]{#1}
\providecommand{\url}[1]{\texttt{#1}}
\expandafter\ifx\csname urlstyle\endcsname\relax
  \providecommand{\doi}[1]{doi: #1}\else
  \providecommand{\doi}{doi: \begingroup \urlstyle{rm}\Url}\fi

\bibitem[Aala \& Van Der~Shaar(2017)Aala and Van Der~Shaar]{guassianprocesses}
A.~Aala and M~Van Der~Shaar.
\newblock Deep multi-task gaussian processes for survival analysis with
  competing risks.
\newblock In \emph{Advances in Neural Information Processing Systems},
  volume~30. Curran Associates, Inc., 2017.

\bibitem[Aalen \& Johansen(1978)Aalen and Johansen]{aalen_empirical_1978}
Odd~O. Aalen and SOren Johansen.
\newblock An {Empirical} {Transition} {Matrix} for {Non}-{Homogeneous} {Markov}
  {Chains} {Based} on {Censored} {Observations}.
\newblock \emph{Scandinavian Journal of Statistics}, 5\penalty0 (3):\penalty0
  141--150, 1978.
\newblock ISSN 0303-6898.
\newblock Publisher: [Board of the Foundation of the Scandinavian Journal of
  Statistics, Wiley].

\bibitem[Alberge et~al.(2025)Alberge, Maladière, Grisel, Abécassis, and
  Varoquaux]{alberge_survival_2025}
Julie Alberge, Vincent Maladière, Olivier Grisel, Judith Abécassis, and Gaël
  Varoquaux.
\newblock Survival {Models}: {Proper} {Scoring} {Rule} and {Stochastic}
  {Optimization} with {Competing} {Risks}.
\newblock \emph{AIStats}, 2025.

\bibitem[Alberge et~al.(2026)Alberge, Haugomat, Varoquaux, and
  Abécassis]{alberge2026calibrationsurvivalmodelscompeting}
Julie Alberge, Tristan Haugomat, Gaël Varoquaux, and Judith Abécassis.
\newblock On the calibration of survival models with competing risks.
\newblock \emph{AISTATS}, 2026.

\bibitem[Antolini et~al.(2005)Antolini, Boracchi, and
  Biganzoli]{antolini_timedependent_2005}
Laura Antolini, Patrizia Boracchi, and Elia Biganzoli.
\newblock A time‐dependent discrimination index for survival data.
\newblock \emph{Statistics in Medicine}, 24\penalty0 (24):\penalty0 3927--3944,
  December 2005.
\newblock ISSN 0277-6715, 1097-0258.

\bibitem[Arslan \& Zeydan(2019)Arslan and Zeydan]{hdfail}
S.~S. Arslan and E.~Zeydan.
\newblock Hard drive failure data.
\newblock Kaggle, 2019.
\newblock URL
  \url{https://www.kaggle.com/datasets/ezeydan/hard-drive-failure-data}.

\bibitem[Bellot \& Schaar(2018)Bellot and Schaar]{bellot_tree-based_2018}
Alexis Bellot and Mihaela Schaar.
\newblock Tree-based {Bayesian} {Mixture} {Model} for {Competing} {Risks}.
\newblock In \emph{Proceedings of the {Twenty}-{First} {International}
  {Conference} on {Artificial} {Intelligence} and {Statistics}}, pp.\
  910--918. PMLR, March 2018.
\newblock ISSN: 2640-3498.

\bibitem[Blanche et~al.(2013)Blanche, Dartigues, and
  Jacqmin-Gadda]{blanche_estimating_2013}
Paul Blanche, Jean-Fran{\c{c}}ois Dartigues, and H{\'e}l{\`e}ne Jacqmin-Gadda.
\newblock Estimating and comparing time-dependent areas under receiver
  operating characteristic curves for censored event times with competing
  risks.
\newblock \emph{Statistics in Medicine}, 2013.

\bibitem[Bolós~Sierra et~al.(2026)Bolós~Sierra, McLean, Friedrich-Welz, Hall,
  and Vallejos]{sierra2026reproducible}
Bego{\~n}a Bolós~Sierra, Colin McLean, Sarah Friedrich-Welz, Peter~S. Hall,
  and Catalina~A. Vallejos.
\newblock A reproducible and extensible framework for benchmarking competing
  risks survival models.
\newblock \emph{arXiv preprint arXiv:2608.00271}, 2026.

\bibitem[Burk et~al.(2026)Burk, Zobolas, Bischl, Bender, Wright, and
  Sonabend]{burk2026neutral}
Lukas Burk, John Zobolas, Bernd Bischl, Andreas Bender, Marvin~N. Wright, and
  Raphael Sonabend.
\newblock A large-scale neutral comparison study of survival models on
  low-dimensional data.
\newblock \emph{arXiv preprint arXiv:2406.04098}, 2026.

\bibitem[Cox(1972)]{cox1972regression}
David~R Cox.
\newblock Regression models and life-tables.
\newblock \emph{Journal of the Royal Statistical Society: Series B
  (Methodological)}, 34\penalty0 (2):\penalty0 187--202, 1972.

\bibitem[D'Agostino \& Nam(2003)D'Agostino and Nam]{dagostino_evaluation_2003}
R.B. D'Agostino and Byung-Ho Nam.
\newblock Evaluation of the {Performance} of {Survival} {Analysis} {Models}:
  {Discrimination} and {Calibration} {Measures}.
\newblock In \emph{Handbook of {Statistics}}, volume~23, pp.\  1--25. Elsevier,
  2003.
\newblock ISBN 978-0-444-50079-3.

\bibitem[Danks \& Yau(2022)Danks and Yau]{danks2022derivative}
Dominic Danks and Christopher Yau.
\newblock Derivative-based neural modelling of cumulative distribution
  functions for survival analysis.
\newblock In \emph{International Conference on Artificial Intelligence and
  Statistics (AISTATS)}. PMLR, 2022.

\bibitem[Dispenzieri et~al.(2010)Dispenzieri, Katzmann, Kyle, Larson, Benson,
  Mullaney, and et~al.]{dispenzieri2010flchain}
Angela Dispenzieri, Jerry~A Katzmann, Robert~A Kyle, Dirk~R Larson, J~Benson,
  C~Mullaney, and et~al.
\newblock Detection of monoclonal gammopathy of undetermined significance
  (mgus) in serum free light chain assays.
\newblock \emph{Blood}, 116\penalty0 (12):\penalty0 2117--2120, 2010.

\bibitem[Djangang et~al.(2025)Djangang, Han, and
  Sanyal]{djangang2025comparative}
Paul~M Djangang, Summer~S Han, and Nilotpal Sanyal.
\newblock Comparative review of modern competing risk methods in
  high-dimensional settings.
\newblock \emph{arXiv preprint arXiv:2503.12824}, 2025.

\bibitem[Drysdale(2022)]{drysdale2022}
Erik Drysdale.
\newblock {SurvSet}: An open-source time-to-event dataset repository.
\newblock \emph{arXiv preprint arXiv:2203.03094}, 2022.

\bibitem[Erickson et~al.(2025)Erickson, Purucker, Tschalzev, Holzm{\"u}ller,
  Mutalik~Desai, Salinas, and Hutter]{tabarena2024}
Nick Erickson, Lennart Purucker, Andrej Tschalzev, David Holzm{\"u}ller,
  Prateek Mutalik~Desai, David Salinas, and Frank Hutter.
\newblock {TabArena}: A living benchmark for machine learning on tabular data.
\newblock In \emph{Advances in Neural Information Processing Systems 38,
  Datasets and Benchmarks Track}, 2025.
\newblock arXiv:2506.16791.

\bibitem[Fine \& Gray(1999)Fine and Gray]{fine_proportional_1999}
Jason~P. Fine and Robert~J. Gray.
\newblock A {Proportional} {Hazards} {Model} for the {Subdistribution} of a
  {Competing} {Risk}.
\newblock \emph{Journal of the American Statistical Association}, 94\penalty0
  (446):\penalty0 496--509, June 1999.
\newblock ISSN 0162-1459, 1537-274X.

\bibitem[Fleming \& Harrington(1991)Fleming and
  Harrington]{fleming1991counting}
Thomas~R Fleming and David~P Harrington.
\newblock \emph{Counting Processes and Survival Analysis}.
\newblock Wiley, New York, 1991.
\newblock ISBN 978-0471522188.

\bibitem[Gneiting \& Raftery(2007)Gneiting and Raftery]{gneiting_strictly_2007}
Tilmann Gneiting and Adrian~E Raftery.
\newblock Strictly {Proper} {Scoring} {Rules}, {Prediction}, and {Estimation}.
\newblock \emph{Journal of the American Statistical Association}, 102\penalty0
  (477):\penalty0 359--378, March 2007.
\newblock ISSN 0162-1459, 1537-274X.

\bibitem[Gorishniy et~al.(2025)Gorishniy, Kotelnikov, and
  Babenko]{gorishniy2025tabm}
Yury Gorishniy, Akim Kotelnikov, and Artem Babenko.
\newblock {TabM}: Advancing tabular deep learning with parameter-efficient
  ensembling.
\newblock In \emph{International Conference on Learning Representations
  (ICLR)}, 2025.

\bibitem[Graf et~al.(1999)Graf, Schmoor, Sauerbrei, and
  Schumacher]{graf_assessment_1999}
Erika Graf, Claudia Schmoor, Willi Sauerbrei, and Martin Schumacher.
\newblock Assessment and comparison of prognostic classification schemes for
  survival data.
\newblock \emph{Statistics in Medicine}, 18\penalty0 (17-18):\penalty0
  2529--2545, September 1999.
\newblock ISSN 0277-6715, 1097-0258.

\bibitem[Guan(2021)]{guan2021properscoringrulevalidation}
Zoe Guan.
\newblock A proper scoring rule for validation of competing risks models.
\newblock \emph{arXiv preprint arXiv:2104.01000}, 2021.

\bibitem[Haider et~al.(2020)Haider, Hoehn, Davis, and
  Greiner]{haider2020effective}
Humza Haider, Bret Hoehn, Sarah Davis, and Russell Greiner.
\newblock {Effective ways to build and evaluate individual survival
  distributions}.
\newblock \emph{Journal of Machine Learning Research}, 21\penalty0
  (85):\penalty0 1--63, 2020.

\bibitem[Harrell~Jr et~al.(1996)Harrell~Jr, Lee, and
  Mark]{harrell1996multivariable}
Frank~E Harrell~Jr, Kerry~L Lee, and Daniel~B Mark.
\newblock Multivariable prognostic models: issues in developing models,
  evaluating assumptions and adequacy, and measuring and reducing errors.
\newblock \emph{Statistics in medicine}, 15\penalty0 (4):\penalty0 361--387,
  1996.

\bibitem[He et~al.(2016)He, Zhang, Ren, and Sun]{resnet}
Kaiming He, Xiangyu Zhang, Shaoqing Ren, and Jian Sun.
\newblock Deep residual learning for image recognition.
\newblock In \emph{Proceedings of the IEEE conference on computer vision and
  pattern recognition}, pp.\  770--778, 2016.

\bibitem[Holzm{\"u}ller et~al.(2024)Holzm{\"u}ller, Grinsztajn, and
  Steinwart]{holzmuller2024better}
David Holzm{\"u}ller, L{\'e}o Grinsztajn, and Ingo Steinwart.
\newblock Better by default: Strong pre-tuned mlps and boosted trees on tabular
  data.
\newblock \emph{Advances in Neural Information Processing Systems},
  37:\penalty0 26577--26658, 2024.

\bibitem[Howard et~al.(2017)Howard, Zhu, Chen, Kalenichenko, Wang, Weyand,
  Andreetto, and Adam]{mobilenet}
Andrew~G. Howard, Menglong Zhu, Bo~Chen, Dmitry Kalenichenko, Weijun Wang,
  Tobias Weyand, Marco Andreetto, and Hartwig Adam.
\newblock Mobilenets: Efficient convolutional neural networks for mobile vision
  applications.
\newblock \emph{arXiv preprint arXiv:1704.04861}, 2017.

\bibitem[Huang et~al.(2017)Huang, Liu, Van Der~Maaten, and
  Weinberger]{densenet}
Gao Huang, Zhuang Liu, Laurens Van Der~Maaten, and Kilian~Q Weinberger.
\newblock Densely connected convolutional networks.
\newblock In \emph{2017 IEEE conference on computer vision and pattern
  recognition (CVPR)}, pp.\  2261--2269. IEEE, 2017.

\bibitem[Ishwaran et~al.(2014)Ishwaran, Gerds, Kogalur, Moore, Gange, and
  Lau]{ishwaran_random_2014}
Hemant Ishwaran, Thomas~A. Gerds, Udaya~B. Kogalur, Richard~D. Moore,
  Stephen~J. Gange, and Bryan~M. Lau.
\newblock Random survival forests for competing risks.
\newblock \emph{Biostatistics}, 15\penalty0 (4):\penalty0 757--773, October
  2014.
\newblock ISSN 1465-4644.

\bibitem[Jeanselme et~al.(2023)Jeanselme, Yoon, Tom, and
  Barrett]{jeanselme2023neuralfinegraymonotonicneural}
Vincent Jeanselme, Chang~Ho Yoon, Brian Tom, and Jessica Barrett.
\newblock Neural fine-gray: Monotonic neural networks for competing risks.
\newblock In \emph{Conference on Health, Inference, and Learning}, pp.\
  379--392. PMLR, 2023.

\bibitem[Jeanselme et~al.(2024)Jeanselme, Agarwal, and
  Wang]{jeanselme2024review}
Vincent Jeanselme, Nikita Agarwal, and Chen Wang.
\newblock Review of language models for survival analysis.
\newblock In \emph{AAAI 2024 Spring Symposium on Clinical Foundation Models},
  2024.

\bibitem[Johnson et~al.(2016)Johnson, Pollard, and
  Mark]{PhysioNet-mimiciii-1.4}
A.~Johnson, T.~Pollard, and R.~Mark.
\newblock {MIMIC-III Clinical Database (version 1.4)}.
\newblock \emph{PhysioNet}, 2016.

\bibitem[Johnson et~al.(2024)Johnson, Bulgarelli, Pollard, Gow, Moody, Horng,
  Celi, and Mark]{PhysioNet-mimiciv-3.1}
A.~Johnson, L.~Bulgarelli, T.~Pollard, B.~Gow, B.~Moody, S.~Horng, L.~A. Celi,
  and R.~Mark.
\newblock {MIMIC-IV (version 3.1)}.
\newblock \emph{PhysioNet}, 2024.

\bibitem[Kandoth et~al.(2013)Kandoth, McLellan, Vandin, Ye, Niu, Lu, Xie,
  Zhang, McMichael, Wyczalkowski, et~al.]{kandoth2013mutational}
Cyriac Kandoth, Michael~D McLellan, Fabio Vandin, Kai Ye, Beifang Niu, Charles
  Lu, Mingchao Xie, Qunyuan Zhang, Joshua~F McMichael, Matthew~A Wyczalkowski,
  et~al.
\newblock Mutational landscape and significance across 12 major cancer types.
\newblock \emph{Nature}, 502\penalty0 (7471):\penalty0 333--339, 2013.

\bibitem[Kaplan \& Meier(1958)Kaplan and Meier]{kaplan_nonparametric_1958}
E.~L. Kaplan and Paul Meier.
\newblock Nonparametric {Estimation} from {Incomplete} {Observations}.
\newblock \emph{Journal of the American Statistical Association}, 53\penalty0
  (282):\penalty0 457--481, June 1958.
\newblock ISSN 0162-1459, 1537-274X.

\bibitem[Katzman et~al.(2018)Katzman, Shaham, Cloninger, Bates, Jiang, and
  Kluger]{katzman2018deepsurv}
Jared~L Katzman, Uri Shaham, Alexander Cloninger, Jonathan Bates, Tingting
  Jiang, and Yuval Kluger.
\newblock Deepsurv: personalized treatment recommender system using a cox
  proportional hazards deep neural network.
\newblock \emph{BMC medical research methodology}, 18\penalty0 (1):\penalty0
  24, 2018.

\bibitem[Kopper et~al.(2024)Kopper, R{\"u}gamer, Sonabend, Bischl, and
  Bender]{kopper2024training}
Philipp Kopper, David R{\"u}gamer, Raphael Sonabend, Bernd Bischl, and Andreas
  Bender.
\newblock On training survival models with scoring rules.
\newblock \emph{arXiv preprint arXiv:2403.13150}, 2024.

\bibitem[Krizhevsky et~al.(2012)Krizhevsky, Sutskever, and
  Hinton]{krizhevsky2012imagenet}
Alex Krizhevsky, Ilya Sutskever, and Geoffrey~E Hinton.
\newblock Imagenet classification with deep convolutional neural networks.
\newblock In \emph{Advances in neural information processing systems}, pp.\
  1097--1105. 2012.

\bibitem[Kvamme et~al.(2019)Kvamme, Borgan, and Scheel]{kvamme2019pycox}
H{\aa}vard Kvamme, {\O}rnulf Borgan, and Ida Scheel.
\newblock {Time-to-Event Prediction with Neural Networks and Cox Regression}.
\newblock \emph{Journal of Machine Learning Research}, 20:\penalty0 1--30,
  2019.

\bibitem[Kvamme \& Borgan(2019)Kvamme and Borgan]{kvamme_continuous_2019}
Håvard Kvamme and Ornulf Borgan.
\newblock Continuous and {Discrete}-{Time} {Survival} {Prediction} with
  {Neural} {Networks}, October 2019.
\newblock arXiv:1910.06724 [cs, stat].

\bibitem[Lee et~al.(2018)Lee, Zame, Yoon, and Van Der~Schaar]{lee_deephit_2018}
Changhee Lee, William Zame, Jinsung Yoon, and Mihaela Van Der~Schaar.
\newblock {DeepHit}: {A} {Deep} {Learning} {Approach} to {Survival} {Analysis}
  {With} {Competing} {Risks}.
\newblock \emph{Proceedings of the AAAI Conference on Artificial Intelligence},
  32\penalty0 (1), April 2018.
\newblock ISSN 2374-3468, 2159-5399.

\bibitem[Lillelund et~al.(2026)Lillelund, Qi, Greiner, and
  Pedersen]{lillelund2025stopchasingcindexevaluate}
Christian~Marius Lillelund, Shi-ang Qi, Russell Greiner, and Christian~Fischer
  Pedersen.
\newblock Position: Stop chasing the c-index when evaluating survival analysis
  models.
\newblock In \emph{Proceedings of the 43rd International Conference on Machine
  Learning (ICML), Position Papers}, 2026.
\newblock arXiv:2506.02075.

\bibitem[Liu et~al.(2024)Liu, Ji, Zhang, and Ye]{liu10385148}
Pei Liu, Luping Ji, Xinyu Zhang, and Feng Ye.
\newblock Pseudo-bag mixup augmentation for multiple instance learning-based
  whole slide image classification.
\newblock \emph{IEEE Transactions on Medical Imaging}, 43\penalty0
  (5):\penalty0 1841--1852, 2024.

\bibitem[Lu et~al.(2021)Lu, Williamson, Chen, Chen, Barbieri, and
  Mahmood]{lu2021data}
Ming~Y Lu, Drew~FK Williamson, Tiffany~Y Chen, Richard~J Chen, Matteo Barbieri,
  and Faisal Mahmood.
\newblock Data-efficient and weakly supervised computational pathology on
  whole-slide images.
\newblock \emph{Nature biomedical engineering}, 5\penalty0 (6):\penalty0
  555--570, 2021.

\bibitem[Mangravite(2013)]{metabric}
Lara Mangravite.
\newblock Metabric data for use in independent research, 2013.

\bibitem[Mesinovic \& Zhu(2025)Mesinovic and
  Zhu]{mesinovic2025survbenchstandardisedpreprocessingpipeline}
Munib Mesinovic and Tingting Zhu.
\newblock Survbench: A standardised preprocessing pipeline for multi-modal
  electronic health record survival analysis.
\newblock \emph{arXiv 2511.11935}, 2025.

\bibitem[Nagpal et~al.(2021)Nagpal, Li, and Dubrawski]{nagpal_deep_2021}
Chirag Nagpal, Xinyu~Rachel Li, and Artur Dubrawski.
\newblock Deep survival machines: Fully parametric survival regression and
  representation learning for censored data with competing risks.
\newblock \emph{IEEE Journal of Biomedical and Health Informatics}, 25\penalty0
  (8):\penalty0 3163--3175, 2021.
\newblock \doi{10.1109/JBHI.2021.3052441}.

\bibitem[Nagpal et~al.(2022)Nagpal, Potosnak, and Dubrawski]{nagpal2022auton}
Chirag Nagpal, William Potosnak, and Aleksandar Dubrawski.
\newblock {auton-survival: An open-source package for regression,
  counter-factual estimation, evaluation and phenotyping with censored
  time-to-event data}.
\newblock In \emph{Machine Learning for Healthcare Conference}, pp.\  585--608,
  2022.

\bibitem[Network et~al.(2013)Network, Weinstein, Collisson, Mills, Mills~Shaw,
  Ozenberger, Ellrott, Shmulevich, Sander, and Stuart]{TCGA2013}
Cancer Genome Atlas~Research Network, John~N. Weinstein, Eric~A. Collisson,
  Gordon~B. Mills, Kenna~R. Mills~Shaw, Brad~A. Ozenberger, Kyle Ellrott, Ilya
  Shmulevich, Chris Sander, and Joshua~M. Stuart.
\newblock The cancer genome atlas pan-cancer analysis project.
\newblock \emph{Nature Genetics}, 45\penalty0 (10):\penalty0 1113--1120, 2013.

\bibitem[Noroozizadeh et~al.(2026)Noroozizadeh, Shen, Weiss, and
  Chen]{noroozizadeh2026survhtebenchbenchmarkheterogeneoustreatment}
Shahriar Noroozizadeh, Xiaobin Shen, Jeremy~C. Weiss, and George~H. Chen.
\newblock {SurvHTE-Bench}: A benchmark for heterogeneous treatment effect
  estimation in survival analysis.
\newblock In \emph{International Conference on Learning Representations
  (ICLR)}, 2026.
\newblock arXiv:2603.05483.

\bibitem[Pedregosa et~al.(2011)Pedregosa, Varoquaux, Gramfort, Michel, Thirion,
  Grisel, Blondel, Prettenhofer, Weiss, Dubourg, Vanderplas, Passos,
  Cournapeau, Brucher, Perrot, and Duchesnay]{scikit-learn}
F.~Pedregosa, G.~Varoquaux, A.~Gramfort, V.~Michel, B.~Thirion, O.~Grisel,
  M.~Blondel, P.~Prettenhofer, R.~Weiss, V.~Dubourg, J.~Vanderplas, A.~Passos,
  D.~Cournapeau, M.~Brucher, M.~Perrot, and E.~Duchesnay.
\newblock Scikit-learn: Machine learning in {P}ython.
\newblock \emph{Journal of Machine Learning Research}, 12:\penalty0 2825--2830,
  2011.

\bibitem[Pintilie(2006)]{pintilie2006competing}
Melania Pintilie.
\newblock \emph{Competing Risks: A Practical Perspective}.
\newblock Wiley, Chichester, England, 2006.
\newblock ISBN 978-0-470-87068-6.

\bibitem[Pollard et~al.(2018)Pollard, Johnson, Raffa, Celi, Mark, and
  Badawi]{pollard_eicu_2018}
Tom~J. Pollard, Alistair E.~W. Johnson, Jesse~D. Raffa, Leo~A. Celi, Roger~G.
  Mark, and Omar Badawi.
\newblock The {eICU} {Collaborative} {Research} {Database}, a freely available
  multi-center database for critical care research.
\newblock \emph{Scientific Data}, 5\penalty0 (1):\penalty0 180178, September
  2018.
\newblock ISSN 2052-4463.

\bibitem[P{\"o}lsterl(2020)]{sksurv}
Sebastian P{\"o}lsterl.
\newblock scikit-survival: A library for time-to-event analysis built on top of
  scikit-learn.
\newblock \emph{Journal of Machine Learning Research}, 21\penalty0
  (212):\penalty0 1--6, 2020.

\bibitem[Qi et~al.(2024{\natexlab{a}})Qi, Sun, and
  Greiner]{qi_survivaleval_2024}
Shi-ang Qi, Weijie Sun, and Russell Greiner.
\newblock {SurvivalEVAL}: {A} {Comprehensive} {Open}-{Source} {Python}
  {Package} for {Evaluating} {Individual} {Survival} {Distributions}.
\newblock \emph{Proceedings of the AAAI Symposium Series}, 2\penalty0
  (1):\penalty0 453--457, 2024{\natexlab{a}}.

\bibitem[Qi et~al.(2024{\natexlab{b}})Qi, Yu, and Greiner]{qi_toward_2025}
Shi-ang Qi, Yakun Yu, and Russell Greiner.
\newblock Toward conditional distribution calibration in survival prediction.
\newblock In \emph{Advances in Neural Information Processing Systems 38
  (NeurIPS)}, 2024{\natexlab{b}}.
\newblock arXiv:2410.20579.

\bibitem[Ries \& Seth(2025)Ries and
  Seth]{ries2025tracertransformerbasedcompetingrisk}
Maxmillan Ries and Sohan Seth.
\newblock Tracer: Transformer-based competing risk analysis with longitudinal
  covariates, 2025.
\newblock URL \url{https://arxiv.org/abs/2512.18129}.

\bibitem[Rindt et~al.(2022)Rindt, Hu, Steinsaltz, and
  Sejdinovic]{rindt_survival_2022}
David Rindt, Robert Hu, David Steinsaltz, and Dino Sejdinovic.
\newblock Survival regression with proper scoring rules and monotonic neural
  networks.
\newblock In \emph{Proceedings of the 25th International Conference on
  Artificial Intelligence and Statistics (AISTATS)}, volume 151 of
  \emph{Proceedings of Machine Learning Research}. PMLR, 2022.

\bibitem[Rueda et~al.(2019)Rueda, Sammut, Seoane, Chin, Caswell-Jin, Callari,
  Batra, Pereira, Bruna, Ali, et~al.]{rueda2019dynamics}
Oscar~M Rueda, Stephen-John Sammut, Jose~A Seoane, Suet-Feung Chin, Jennifer~L
  Caswell-Jin, Maurizio Callari, Rajbir Batra, Bernard Pereira, Alejandra
  Bruna, H~Raza Ali, et~al.
\newblock Dynamics of breast-cancer relapse reveal late-recurring er-positive
  genomic subgroups.
\newblock \emph{Nature}, 567\penalty0 (7748):\penalty0 399--404, 2019.

\bibitem[SEER(2025)]{seer_dataset}
SEER.
\newblock {SEER database}.
\newblock \url{https://seer.cancer.gov/causespecific}, 2025.

\bibitem[{TorchVision maintainers and contributors}(2016)]{torchvision2016}
{TorchVision maintainers and contributors}.
\newblock {TorchVision}: {PyTorch}'s computer vision library.
\newblock \url{https://github.com/pytorch/vision}, 2016.

\bibitem[Tsao \& Vasan(2015)Tsao and Vasan]{Tsao2015}
Connie~W Tsao and Ramachandran~S Vasan.
\newblock The framingham heart study: past, present and future.
\newblock \emph{International Journal of Epidemiology}, 44\penalty0
  (6):\penalty0 1763--1766, 12 2015.

\bibitem[Uno et~al.(2007)Uno, Cai, Tian, and Wei]{uno_evaluating_2007}
Hajime Uno, Tianxi Cai, Lu~Tian, and L.~J. Wei.
\newblock Evaluating prediction rules for t-year survivors with censored
  regression models.
\newblock \emph{Journal of the American Statistical Association}, 102\penalty0
  (478):\penalty0 527--537, 2007.

\bibitem[Uno et~al.(2011)Uno, Cai, Pencina, D'Agostino, and
  Wei]{uno_cstatistics_2011}
Hajime Uno, Tianxi Cai, Michael~J. Pencina, Ralph~B. D'Agostino, and L.~J. Wei.
\newblock On the {C}‐statistics for evaluating overall adequacy of risk
  prediction procedures with censored survival data.
\newblock \emph{Statistics in Medicine}, 30\penalty0 (10):\penalty0 1105--1117,
  May 2011.
\newblock ISSN 0277-6715, 1097-0258.

\bibitem[Van~Belle et~al.(2011)Van~Belle, Pelckmans, Van~Huffel, and
  Suykens]{van_belle_support_2011}
Vanya Van~Belle, Kristiaan Pelckmans, Sabine Van~Huffel, and Johan~A.K.
  Suykens.
\newblock Support vector methods for survival analysis: a comparison between
  ranking and regression approaches.
\newblock \emph{Artificial Intelligence in Medicine}, 53\penalty0 (2):\penalty0
  107--118, October 2011.
\newblock ISSN 09333657.

\bibitem[Van~Calster et~al.(2019)Van~Calster, McLernon, Van~Smeden, Wynants,
  Steyerberg, and {Topic Group ‘Evaluating diagnostic tests and prediction
  models’ of the STRATOS initiative}]{van2019calibration}
Ben Van~Calster, David~J McLernon, Maarten Van~Smeden, Laure Wynants, Ewout~W
  Steyerberg, and {Topic Group ‘Evaluating diagnostic tests and prediction
  models’ of the STRATOS initiative}.
\newblock Calibration: the achilles heel of predictive analytics.
\newblock \emph{BMC medicine}, 17\penalty0 (1):\penalty0 230, 2019.

\bibitem[Van~Geloven et~al.(2022)Van~Geloven, Giardiello, Bonneville, Teece,
  Ramspek, Van~Smeden, Snell, Van~Calster, Pohar-Perme, Riley, Putter, and
  Steyerberg]{van_geloven_validation_2022}
Nan Van~Geloven, Daniele Giardiello, Edouard~F Bonneville, Lucy Teece, Chava~L
  Ramspek, Maarten Van~Smeden, Kym I~E Snell, Ben Van~Calster, Maja
  Pohar-Perme, Richard~D Riley, Hein Putter, and Ewout Steyerberg.
\newblock Validation of prediction models in the presence of competing risks: a
  guide through modern methods.
\newblock \emph{BMJ}, pp.\  e069249, May 2022.
\newblock ISSN 1756-1833.

\bibitem[Wang \& Sun(2022)Wang and Sun]{wang_survtrace_2022}
Zifeng Wang and Jimeng Sun.
\newblock {SurvTRACE}: {Transformers} for {Survival} {Analysis} with
  {Competing} {Events}.
\newblock In \emph{Proceedings of the 13th {ACM} {International} {Conference}
  on {Bioinformatics}, {Computational} {Biology} and {Health} {Informatics}},
  pp.\  1--9, August 2022.
\newblock arXiv:2110.00855 [cs, stat].

\bibitem[Wang et~al.(2024)Wang, Lee, Chakraborty, Ning, Liu, Xie, Ong, and
  Liu]{wang2024survivalmodelingusingdeep}
Ziwen Wang, Jin~Wee Lee, Tanujit Chakraborty, Yilin Ning, Mingxuan Liu, Feng
  Xie, Marcus Eng~Hock Ong, and Nan Liu.
\newblock Survival modeling using deep learning, machine learning and
  statistical methods: A comparative analysis for predicting mortality after
  hospital admission.
\newblock \emph{arxiv 2403.06999}, 2024.

\bibitem[Wiegrebe et~al.(2024)Wiegrebe, Kopper, Sonabend, Bischl, and
  Bender]{wiegrebe2024deep}
Simon Wiegrebe, Philipp Kopper, Raphael Sonabend, Bernd Bischl, and Andreas
  Bender.
\newblock Deep learning for survival analysis: a review.
\newblock \emph{Artificial Intelligence Review}, 57\penalty0 (3):\penalty0 65,
  2024.

\bibitem[Wolbers et~al.(2009)Wolbers, Koller, Witteman, and
  Steyerberg]{wolbers_prognostic_2009}
Marcel Wolbers, Michael~T. Koller, Jacqueline C.~M. Witteman, and Ewout~W.
  Steyerberg.
\newblock Prognostic {Models} {With} {Competing} {Risks}: {Methods} and
  {Application} to {Coronary} {Risk} {Prediction}.
\newblock \emph{Epidemiology}, 20\penalty0 (4):\penalty0 555--561, July 2009.
\newblock ISSN 1044-3983.

\bibitem[Yanagisawa(2023)]{yanagisawa_proper_2023}
Hiroki Yanagisawa.
\newblock Proper scoring rules for survival analysis.
\newblock In \emph{Proceedings of the 40th International Conference on Machine
  Learning (ICML)}, volume 202 of \emph{Proceedings of Machine Learning
  Research}. PMLR, 2023.

\end{thebibliography}
\bibliographystyle{iclr2027_conference}

\newpage
\appendix
\counterwithin{figure}{section}
\counterwithin{table}{section}
\onecolumn
\part*{Appendices}

We give detailed appendices to be precise on all definitions and help reproducibility.
\smallskip

\enableappendixtoc
\setcounter{tocdepth}{2}
\appendixtoc

\section{Details on Datasets}\label{sec:datasets}
\subsection{Access and preprocessing}\label{sec:availability}

The following instructions cover the 20 tabular datasets in Table~\ref{tab:datasetsstatistics}. Paths are relative to the \texttt{benchmark/} directory of the code to be released. The dataset-specific loaders are in \texttt{\_dataset.py}, with the PBC preparation in \texttt{auton\_datasets.py}. Each loader returns covariates and a target with columns \texttt{duration} and \texttt{event}; event code 0 denotes right-censoring. For prepared CSV files, obtaining the source data and constructing the cohort are separate steps.

\paragraph{Shared tabular preparation.}
The custom CSV loaders, the competing-risks METABRIC loader, and the SurvSet loader apply \texttt{skrub.TableVectorizer}, separate the target columns from the transformed covariates, and fill remaining missing covariate values with column means. They cast covariates to \texttt{float32} and events to integers. Rows with missing targets are removed where specified below. The pycox loaders and the SEER breast loader use their own column typing, described in their entries. The dataset-specific operations below precede the benchmark split and model fitting.

\paragraph{Framingham \citep{Tsao2015}.}
The local \texttt{datasets/framingham.csv} has the same 4,240 rows and column names as the \texttt{cvdd} teaching dataset documented by \texttt{riskCommunicator}.\footnote{\url{https://search.r-project.org/CRAN/refmans/riskCommunicator/html/cvdd.html}}
To obtain that public teaching dataset, install \texttt{riskCommunicator} from CRAN, load \texttt{data(cvdd, package="riskCommunicator")}, and export it with \texttt{write.csv(cvdd, "framingham.csv")}. This source already restricts observations to the first examination and excludes prevalent coronary heart disease. Our \texttt{load\_framingham} uses \texttt{timeout} as the duration in days, codes CVD as event 1 and death without CVD as event 2, drops \texttt{logpdays}, \texttt{cvd\_dth}, \texttt{bmicat}, and \texttt{CURSMOKE}, and removes missing targets before vectorization. The retained columns are listed in the loader.

\paragraph{METABRIC with competing events \citep{rueda2019dynamics}.}
Open the supplementary information for \citet{rueda2019dynamics}\footnote{\url{https://doi.org/10.1038/s41586-019-1007-8}} and obtain \texttt{41586\_2019\_1007\_MOESM7\_ESM.txt}. Place it at the location specified by \texttt{PATH\_METABRIC}. The loader extracts diagnosis year from \texttt{Date.Of.Diagnosis}, normalizes column names, and retains 16 clinical covariates. It removes rows missing any of \texttt{Death}, \texttt{T}, \texttt{TLR}, \texttt{TDR}, \texttt{LR}, or \texttt{DR}. Event 1 denotes local or distant relapse, with duration initially set to $\min(\texttt{TLR},\texttt{TDR})$. Death is event 2 when \texttt{Death}=1 and \texttt{T} is no greater than that duration; the duration is then replaced by \texttt{T}. Other observations retain event code 0. We then apply the shared tabular preparation.

\paragraph{MIMIC-IV \citep{PhysioNet-mimiciv-3.1}.}
Open the PhysioNet page for version 3.1.\footnote{\url{https://physionet.org/content/mimiciv/3.1/}}
Create a PhysioNet account, obtain credentialed access, complete the required research training, and sign the dataset's data use agreement. Download the hospital \texttt{patients}, \texttt{admissions}, and \texttt{labevents} tables and the ICU \texttt{icustays} table. Construct the cohort using Section~\ref{sec:studymimic} and save the prepared table as \texttt{datasets/mimic\_cohort.csv}. The \texttt{load\_mimic} function removes missing targets and applies the shared preparation to the remaining columns.

\paragraph{MIMIC-III \citep{PhysioNet-mimiciii-1.4}.}
Use the PhysioNet page for version 1.4\footnote{\url{https://physionet.org/content/mimiciii/1.4/}} and complete its access requirements. Download \texttt{PATIENTS}, \texttt{ADMISSIONS}, \texttt{ICUSTAYS}, and \texttt{LABEVENTS}. Apply the cohort procedure described in Section~\ref{sec:studymimic}, using the MIMIC-III schema, and save \texttt{datasets/mimiciii\_cohort.csv}. The \texttt{load\_mimiciii} function removes missing targets and selects diagnosis, age, and the 50 laboratory variables explicitly listed in the code before vectorization and mean imputation.

\paragraph{eICU \citep{pollard_eicu_2018}.}
Request access to version 2.0 on PhysioNet.\footnote{\url{https://physionet.org/content/eicu-crd/2.0/}}
The account and approved training used for MIMIC can be reused, but the eICU data use agreement must also be signed. Download the \texttt{patient} and \texttt{lab} tables and construct the cohort described in Section~\ref{sec:studymimic}. Save \texttt{datasets/eicu\_cohort.csv}. The \texttt{load\_eicu} function removes missing targets and retains gender, admission height and weight, cleaned age, and the laboratory columns listed in the loader before applying the shared preparation.

\paragraph{SEER breast cancer \citep{seer_dataset}.}
Request SEER data access through the National Cancer Institute and follow the instructions for SEER*Stat.\footnote{\url{https://seer.cancer.gov/data/access.html}}
Use the \texttt{hazardous} export tutorial\footnote{\url{https://soda-inria.github.io/hazardous/downloading_seer.html}} to obtain the tab-separated file expected by \texttt{PATH\_SEER}, preserving its 29-column order. We call \texttt{load\_seer} with \texttt{survtrace\_preprocessing=True}. This removes records with the specified unavailable cause-specific classification or death before recommended surgery, recodes selected unknown values, and groups rare categories as implemented in \texttt{hazardous/data/\_seer.py}. We retain 14 categorical and four numerical predictors. Duration is \texttt{Survival months}; events are breast cancer death, heart disease death, and other causes of death, coded 1, 2, and 3, respectively. Alive patients are censored.

\paragraph{SEER digestive cancers \citep{seer_dataset}.}
Use the same SEER access procedure and export the digestive-cancer cohort in SEER*Stat, including survival months, cause of death, and clinical covariates. The benchmark reads the prepared file \texttt{datasets/seer\_cohort.csv} through \texttt{load\_seer\_digestion}, removes missing \texttt{event} or \texttt{duration}, and applies the shared preparation to all remaining columns. Reproducing the raw-data extraction requires the original SEER*Stat session, including the database release, site filters, and event recoding; these are not recorded in this loader.

\paragraph{PBC \citep{fleming1991counting}.}
Obtain \texttt{pbc2.csv} from the \texttt{auton-survival} dataset collection\footnote{\url{https://github.com/autonlab/auton-survival/tree/master/auton_survival/datasets}} and place it in \texttt{datasets/}. Use the benchmark's \texttt{load\_pbc}, which calls the local \texttt{load\_dataset\_auton("PBC")}. The local helper keeps the repeated observations as separate rows, dummy-encodes seven categorical variables, retains seven laboratory variables, and derives age as \texttt{age + years}. Duration is \texttt{years - year}; transplantation is event 1 and death is event 2. It mean-imputes and standardizes the covariates. The outer loader removes missing targets and applies the shared preparation. Thus the sample count refers to observations, not distinct patients.

\paragraph{Loan.}
Register on Kaggle and download the Lending Club archive.\footnote{\url{https://www.kaggle.com/datasets/wordsforthewise/lending-club}}
The benchmark uses a derived \texttt{datasets/loan\_cohort.csv}. The \texttt{load\_loan} function removes duplicate interest-rate and installment columns and \texttt{sub\_grade}, converts the loan term from text to months, and converts employment length to a number, coding ``10+ years'' as 10 and ``< 1 year'' as 0. It renames \texttt{time\_to\_event} to \texttt{duration} and applies the shared preparation. The prepared file also includes four macroeconomic covariates. Its construction script is needed to reproduce the loan-status event mapping, time origin, censoring date, and macroeconomic joins; these operations are not performed by the loader.

\paragraph{TCGA \citep{TCGA2013}.}
Clinical TCGA data can be obtained through cBioPortal.\footnote{\url{https://www.cbioportal.org}}
Our tabular loader expects the prepared \texttt{datasets/tcga.csv}, including the competing-event columns \texttt{DSS\_cr} and \texttt{DSS.time.cr}. It renames these to \texttt{event} and \texttt{duration}, removes missing targets, and excludes patient barcodes, vital status, follow-up and death dates, cause of death, new-tumor-event descriptors, treatment outcome, margin status, and residual tumor. It then applies the shared preparation. A generic cBioPortal download does not establish how the two competing-event columns were derived: the original clinical export and endpoint-construction script are needed to reproduce this step. The results are based on data generated by the TCGA Research Network.\footnote{\url{https://www.cancer.gov/tcga}}

\paragraph{FLCHAIN \citep{dispenzieri2010flchain}.}
Install \texttt{pycox}\footnote{\url{https://github.com/havakv/pycox}} and call \texttt{pycox.datasets.flchain.read\_df(processed=False)}. Our \texttt{get\_split\_flchain} removes rows missing creatinine, drops \texttt{chapter} and \texttt{rownames}, and encodes sex as a male indicator. It treats \texttt{sample.yr} and \texttt{flc.grp} as categorical and casts the other columns to \texttt{float32}. The target uses \texttt{futime} for duration and \texttt{death} for the event indicator.

\paragraph{Hdfail \citep{hdfail}.}
Install \texttt{SurvSet}\footnote{\url{https://github.com/ErikinBC/SurvSet}} and obtain the data with \texttt{SurvLoader().load\_dataset("hdfail")["df"]} from \texttt{SurvSet.data}. The benchmark calls \texttt{load\_survset("hdfail")}: it renames \texttt{time} to \texttt{duration}, drops rows missing a target and removes \texttt{pid}, then vectorizes and mean-imputes the remaining covariates. It retains the supplied failure indicator and follow-up times.

\paragraph{GBSG \citep{katzman2018deepsurv}.}
Obtain the packaged benchmark with \texttt{pycox.datasets.gbsg.read\_df()}. Our \texttt{get\_split\_gbsg} retains its supplied \texttt{event} and \texttt{duration}, treats all seven covariates \texttt{x0}--\texttt{x6} as numerical, and casts them to \texttt{float64}. Use this packaged version to match the 2,232-row benchmark.

\paragraph{METABRIC, single event \citep{metabric}.}
Call \texttt{pycox.datasets.metabric.read\_df()}. The benchmark retains the supplied overall-survival target and nine covariates, casting \texttt{x4}, \texttt{x5}, \texttt{x6}, and \texttt{x7} as categorical and the remaining five as \texttt{float64}. This is the 1,904-row pycox dataset; the relapse and death construction described above belongs to the separate competing-events dataset.

\paragraph{MSK 2021.}
Open the cBioPortal study \texttt{msk\_met\_2021}\footnote{\url{https://www.cbioportal.org/study/summary?id=msk_met_2021}} and download the clinical-data table as \texttt{datasets/msk\_met\_2021\_clinical\_data.tsv}. Use the clinical export with human-readable column names expected by \texttt{load\_msk}. The loader converts \texttt{NA} to missing values, uses overall survival in months as duration, maps \texttt{0:LIVING} to censoring and \texttt{1:DECEASED} to event 1, and removes missing targets. The 14 retained covariates are listed in \texttt{get\_split\_msk\_2021}; they include age at sequencing, cancer type, gene panel, mutation count, and sample characteristics. We then apply the shared preparation.

\paragraph{MSK 2026.}
Download the clinical-data table for cBioPortal study \texttt{msk\_impact\_50k\_2026}\footnote{\url{https://www.cbioportal.org/study/summary?id=msk_impact_50k_2026}} as \texttt{datasets/msk\_impact\_50k\_2026\_clinical\_data.tsv}. We use the same overall-survival coding and missing-target removal as for MSK 2021. The 15 covariates in \texttt{get\_split\_msk\_2026} include age at diagnosis, cancer type, tumor mutation burden, ploidy, purity, genomic alteration fraction, gene panel, ancestry, MSI score, mutation count, coverage, sex, FACETS quality control, and HLA genotype availability. The shared preparation follows this selection.

\paragraph{NWTCO.}
Call \texttt{pycox.datasets.nwtco.read\_df(processed=False)}. The benchmark uses \texttt{edrel} as duration and \texttt{rel} as the relapse indicator. It creates \texttt{instit\_2 = instit - 1}, \texttt{histol\_2 = histol - 1}, and \texttt{study\_4 = study - 3}; it drops the original three columns, \texttt{rownames}, and \texttt{seqno}. Stage is categorical, and the other covariates are cast to \texttt{float32}.

\paragraph{Smarto.}
Use \texttt{SurvLoader().load\_dataset("smarto")["df"]} from \texttt{SurvSet.data}. The benchmark's \texttt{load\_survset("smarto")} preserves the supplied event indicator, renames \texttt{time} to \texttt{duration}, removes missing targets and \texttt{pid}, and applies the shared preparation to the remaining covariates.

\paragraph{SUPPORT \citep{katzman2018deepsurv}.}
Call \texttt{pycox.datasets.support.read\_df()} to obtain the packaged 8,873-row benchmark. We retain its supplied survival target and 14 covariates. The loader casts \texttt{x1}--\texttt{x6} as categorical and \texttt{x0}, \texttt{x7}--\texttt{x13} as \texttt{float64}.

\paragraph{ENS Challenge.}
Create an account on Challenge Data and download the labeled training files for QRT's ``Overall Survival Prediction for patients diagnosed with Myeloid Leukemia.''\footnote{\url{https://challengedata.ens.fr/challenges/162}}
The source provides clinical measurements, molecular records, and survival targets. Our \texttt{load\_challenge\_ens} reads \texttt{datasets/challenge\_ens\_cohort.csv}, removes missing targets, and applies the shared preparation. The prepared table contains center, six blood or marrow measurements, and the derived columns \texttt{num\_mutated\_genes}, \texttt{nb\_of\_chromosom}, and \texttt{sex\_chromosome}. The original preparation script is needed to reproduce the molecular aggregation and cytogenetic parsing that created these columns.

\subsection{Study on MIMIC III-IV, and eICU}\label{sec:studymimic}
For those ICU datasets, we applied the same methodology.

\subsubsection{Cohort Selection and Inclusion Criteria}

We constructed our cohort from the MIMIC-IV database. To ensure a focused analysis on critical care outcomes, we applied the following inclusion criteria:
\begin{itemize}
    \item \textbf{ICU Admission:} Only patients with at least one recorded ICU stay (\texttt{icustays}) were included.
    \item \textbf{Adult Population:} Patients were filtered to include only those $\geq18$ years of age at the time of their anchor year.
    \item \textbf{Index Stay Selection:} We randomly sampled a single hospital admission (\texttt{hadm\_id}) for each unique patient (\texttt{subject\_id}).
\end{itemize}

\subsubsection{Feature Engineering: Aggregated Clinical Labs}

Due to the high dimensionality and irregular sampling of laboratory measurements, we processed the \texttt{labevents} table using a memory-efficient chunking strategy:

\begin{itemize}
    \item \textbf{Dimensionality Reduction:} We identified the top 50 most frequent lab tests to represent the physiological state during the first day of the stay.
    \item \textbf{Aggregation:} For the selected index stay, all lab observations recorded between \texttt{admittime} and \texttt{admittime} + 1 day were aggregated. We computed the mean values for each test, resulting in a fixed-length feature vector per patient.
\end{itemize}

\subsubsection{Target Definition: Competing Risks Framework}

We treat the post-discharge outcome as a competing risks problem. Let $T$ be the time to the first event following hospital discharge. Because MIMIC-IV dates are shifted for de-identification, we calculated all durations in relative days from the index discharge time.

We define three mutually exclusive outcomes:
\begin{itemize}
    \item \textbf{Rehospitalization (Event Type 1):} Defined as the duration between the current index \texttt{admittime} + 1 day and the \texttt{admittime} of the subsequent hospital visit.
    \item \textbf{Death (Event Type 2):} Defined as the duration between the index \texttt{admittime} + 1 day and the date of death (dod) recorded in the patients table.
    \item \textbf{Administrative Censoring (Event Type 0):} To handle the lack of a global end-date, we applied individual-level administrative censoring. A patient is censored at the time of their last known interaction with the hospital system (the maximum discharge time across all their records).
\end{itemize}

\subsection{Statistics summarizing the datasets}
Table~\ref{tab:datasetsstatistics} reports the dataset characteristics after the dataset-specific preparation described in Section~\ref{sec:availability}. For loaders using \texttt{TableVectorizer}, the feature count is measured after vectorization. The benchmark code will be available on GitHub after publication.

\begin{table*}[h!]
    \caption{\textbf{Characteristics of the 20 tabular datasets.} The CR column indicates whether
    the dataset has competing events. The number of features is counted after applying skrub's
    TableVectorizer. The Github column indicates whether the dataset is directly available from
    GitHub; Sec.~\ref{sec:availability} explains how we obtained the others.}
    \label{tab:datasetsstatistics}
    \centering
    \begin{tabular}{cccccc}
    \toprule
    Dataset name & CR & Samples & Nb features & Censoring rate & Github \\
    \midrule
    Framingham  & \checkmark & 4,240 &24 & 58\% & $\sim$\\
    Metabric (competing events)  & \checkmark & 3,237& 73 & 33\% & \xmark\\
    MIMIC IV & \checkmark & 65,366 & 172 & 58\%& \xmark \\
    MIMIC  III  & \checkmark & 38,511& 81 & 54\% & \xmark \\
    SEER Breast Cancer   & \checkmark & 476,746& 18 & 63\% & \xmark \\
    SEER Digestion Cancer & \checkmark & 887,552& 304 & 19\% & \xmark \\
    PBC  & \checkmark & 1,945 & 25 & 55\% & auton-datasets\\
    Loan & \checkmark & 1,652,304 &25&41\% & \xmark \\
    eICU & \checkmark & 106,196& 58 & 95 \% &\xmark \\
    TCGA  & \checkmark & 10,540 & 151 & 70\%&\xmark \\
    \midrule
    Flchain & \xmark & 6,524& 8 & 70\% &pycox\\
    Hdfail  & \xmark& 52,422 & 38 & 95\% &SurvSet\\
    GBSG   & \xmark & 2,232&7&43\% & pycox\\
    Metabric  & \xmark &1,904&9&42\%& pycox\\
    MSK 2021   & \xmark & 25,659&86& 60\% & \xmark\\
    MSK 2026  & \xmark & 40,755&56& 57\% & \xmark \\
    NWTCO  & \xmark &  4,028 & 6 & 85\% &pycox\\
    smarto  & \xmark & 3,873& 39 & 88\% & SurvSet\\
    Support  & \xmark & 8,873& 14& 32\% & pycox\\
    ENS Challenge & \xmark & 3,173&42&50\% & \xmark \\
    \bottomrule
    \end{tabular}
\end{table*}

\section{Theory: Definitions of the evaluation metrics}\label{sec:metrics}

This appendix gives the precise definition of every metric used in Sec.~\ref{sec:crmetrics}. Throughout, the model outputs a predicted CIF $\hat F_k(t\mid\mathbf{x})$ for each cause $k\in\{1,\dots,K\}$ and a predicted overall survival function $\hat S(t\mid\mathbf{x}) = 1 - \sum_{k=1}^K \hat F_k(t\mid\mathbf{x})$. Metrics are computed on a held-out test set $\D_{\rm test} = \{(\mathbf{x}_i, t_i, \delta_i)\}_{i=1}^{n}$, with $\delta_i\in\{0,1,\dots,K\}$ and $\delta_i = 0$ denoting right-censoring. We write $\hat G(u) = \widehat{\P}(C > u)$ for the Kaplan-Meier estimate of the censoring survival function, fitted on $\D_{\rm test}$ by swapping the event indicator ($\delta_i = 0$ becomes the ``event''). This evaluation-time $\hat G$ is the marginal Kaplan-Meier estimator for every model and every metric; it is distinct from the training-time censoring weights $G^\star$ of the CRNLL loss (Eq.~\ref{eqn:full_loss}, Sec.~\ref{sec:ipcwsensitivity}). Time integrals are approximated on a fixed grid $0 = \tau_0 < \tau_1 < \dots < \tau_M = t_{\max}$ of empirical event-time quantiles, shared by all models on a given dataset, with $t_{\max}$ the $99$th percentile of the observed times. All time-dependent scores below are also reported at the $25$th, $50$th and $75$th percentiles of the observed event times (``C-index 25'', etc.).

\subsection{Metrics controlling probabilistic prediction metrics: IBS and INLL}

For cause $k$ and horizon $t$, let $y_{i,k}(t) = \mathbb{1}(t_i \le t,\ \delta_i = k)$ be the binary label ``subject $i$ has experienced cause $k$ by time $t$'', and let the IPCW weight be
\begin{equation}
\hat w_i(t) \;=\; \frac{\mathbb{1}(t_i \le t,\ \delta_i \neq 0)}{\hat G(t_i)} \;+\; \frac{\mathbb{1}(t_i > t)}{\hat G(t)} .
\end{equation}
Subjects censored before $t$ ($t_i \le t,\ \delta_i = 0$) receive weight $0$; the weights of the others compensate for this missing mass \citep{graf_assessment_1999}.

\begin{definition}[Integrated Brier Score (IBS)]\label{def:ibs}
The time-dependent Brier score and its integrated, cause-averaged version are
\begin{align}
\mathrm{BS}_k(t) &= \frac{1}{n}\sum_{i=1}^{n} \hat w_i(t)\,\big(y_{i,k}(t) - \hat F_k(t\mid\mathbf{x}_i)\big)^2 , \\
\mathrm{IBS}_k   &= \frac{1}{t_{\max}}\int_0^{t_{\max}} \mathrm{BS}_k(t)\,\d t , \qquad
\mathrm{IBS} = \frac{1}{K}\sum_{k=1}^{K}\mathrm{IBS}_k .
\end{align}
\end{definition}

\begin{definition}[Integrated Negative Log-Likelihood (INLL)]\label{def:inll}
INLL replaces the squared error of Def.~\ref{def:ibs} by the Bernoulli log-loss, keeping the same weights $\hat w_i(t)$ and the same time integration:
\begin{align}
\mathrm{NLL}_k(t) &= -\frac{1}{n}\sum_{i=1}^{n}\hat w_i(t)\Big[ y_{i,k}(t)\log \hat F_k(t\mid\mathbf{x}_i) + \big(1 - y_{i,k}(t)\big)\log\big(1 - \hat F_k(t\mid\mathbf{x}_i)\big)\Big], \\
\mathrm{INLL}_k &= \frac{1}{t_{\max}}\int_0^{t_{\max}} \mathrm{NLL}_k(t)\,\d t , \qquad
\mathrm{INLL} = \frac{1}{K}\sum_{k=1}^{K}\mathrm{INLL}_k .
\end{align}
Predicted probabilities are clipped to $[\varepsilon, 1-\varepsilon]$ with $\varepsilon = 10^{-8}$ before taking logs.
\end{definition}

\paragraph{Strictly proper scores} These two metrics are \emph{strictly proper} scoring rules:
Both are minimized (in expectation) \emph{if and only if} $\hat F_k = F_k$ for all $k$, under the assumption that $\hat G$ is correctly specified \citep{gneiting_strictly_2007, rindt_survival_2022}. They should be read as lower is better.

\subsection{Discrimination metrics: C-index and time-dependent AUC}

\begin{definition}[Time-dependent competing-risks C-index]\label{def:cindex}
Following \cite{wolbers_prognostic_2009, antolini_timedependent_2005}, for cause $k$ and horizon $t$ we order pairs $(i,j)$ by the predicted incidence $\hat F_k(t\mid\mathbf{x})$. A pair is \emph{comparable} for cause $k$ if $i$ has experienced cause $k$ by $t$ and $j$ is known to be at risk of a later or competing cause-$k$ event:
\begin{equation}
A^k_{ij} = \mathbb{1}(\delta_i = k,\ t_i \le t)\;\Big[\mathbb{1}(t_j > t_i) \;+\; \mathbb{1}\big(t_j \le t_i,\ \delta_j \notin \{0, k\}\big)\Big].
\end{equation}
The pair is \emph{concordant} if $\hat F_k(t\mid\mathbf{x}_i) > \hat F_k(t\mid\mathbf{x}_j)$, and
\begin{equation}
C_k(t) = \frac{\sum_{i\neq j} A^k_{ij}\,\Big[\mathbb{1}\big(\hat F_k(t\mid\mathbf{x}_i) > \hat F_k(t\mid\mathbf{x}_j)\big) + \tfrac12\,\mathbb{1}\big(\hat F_k(t\mid\mathbf{x}_i) = \hat F_k(t\mid\mathbf{x}_j)\big)\Big]}{\sum_{i\neq j} A^k_{ij}} .
\end{equation}
Comparable pairs are IPCW-weighted by $1/\hat G(t_i)$ to correct for censoring \citep{uno_cstatistics_2011}. $C_k(t) = 0.5$ is random ranking, $C_k(t) = 1$ is perfect ranking; higher is better.
\end{definition}

\begin{definition}[Time-dependent (cumulative/dynamic) AUC]\label{def:auc}
Following \cite{uno_evaluating_2007, blanche_estimating_2013}, at horizon $t$ the \emph{cases} for cause $k$ are $\{i : t_i \le t,\ \delta_i = k\}$ and the \emph{controls} are $\{j : t_j > t\}$. With IPCW weights $\omega_i = 1/\hat G(t_i)$ on the cases,
\begin{equation}
\widehat{\mathrm{AUC}}_k(t) =
\frac{\displaystyle\sum_{i}\sum_{j} \omega_i\,\mathbb{1}(t_i \le t,\ \delta_i = k)\,\mathbb{1}(t_j > t)\,\Big[\mathbb{1}\big(\hat F_k(t\mid\mathbf{x}_i) > \hat F_k(t\mid\mathbf{x}_j)\big) + \tfrac12\mathbb{1}(\text{tie})\Big]}
{\Big(\sum_i \omega_i\,\mathbb{1}(t_i \le t,\ \delta_i = k)\Big)\Big(\sum_j \mathbb{1}(t_j > t)\Big)} .
\end{equation}
It estimates $\P\big(\hat F_k(t\mid\mathbf{X}_i) > \hat F_k(t\mid\mathbf{X}_j) \,\big|\, T^\star_i \le t, \Delta^\star_i = k,\ T^\star_j > t\big)$. Higher is better.
\end{definition}

Both discrimination metrics are invariant to any strictly monotone transformation of the predicted incidences, hence insensitive to calibration.

\subsection{Calibration metrics}

We use the two competing-risks calibration metrics of \cite{alberge2026calibrationsurvivalmodelscompeting}, both computed with $\alpha = 2$ on a held-out calibration split $\mathcal{D}_{cal}$:
\begin{description}
\item[AJ-calibration] (plug-in / Aalen-Johansen calibration): the time-integrated absolute gap between the average predicted CIF $\frac{1}{|\mathcal{D}_{cal}|}\sum_i \hat F_k(\tau\mid\mathbf{x}_i)$ and a marginal Aalen--Johansen estimate \citep{aalen_empirical_1978} of $\P(T^\star \le \tau, \Delta^\star = k)$ on the same split, summed over causes. It targets marginal (population-level) calibration. Full definition in Def.~\ref{def:ajcal}.
\textbf{AJ calibration.} We compute the PI (plug-in) calibration with the Aalen--Johansen estimator \citep{aalen_empirical_1978} as the PI estimator, and $\alpha = 2$. As the Aalen--Johansen estimator is marginal, it predicts, for a given time $\tau$ and a given event $k, \text{AJ}_k(\tau) \approx \mathbb{P}(T^\star \leq \tau, \Delta^\star =k ) $. The metric is computed onto a calibration set called $\mathcal{D}_{cal}$, and each sample $x_i \in \mathcal{X}_{cal}$.

\begin{definition}[Plug-in calibration]\label{def:ajcal}
    \begin{flalign*}
        \text{\footnotesize{Plug-in-${\rm cal}_{K}^\alpha$-calibration:}} &&  {{\rm PI}^\alpha_{cal}} = \sum_{k=1}^K  {{\rm PI}^\alpha_k} &&
    \end{flalign*}
    with
    \begin{flalign*}
        \text{\footnotesize{Plug-in-$k^\alpha$ calibration:}} && {{\rm PI}^\alpha_k} = \left(\int_0^{t_{max}} {\rm PI}_k(\tau)^\alpha \d \tau \right)^{1/\alpha} &&
    \end{flalign*}
    where
    \begin{flalign*}
        \text{\parbox{\linewidth}{\footnotesize{Plug-in-${\rm cal}_{k}(\tau)$ calibration:}}}\\[-.1em] {{\rm PI}_k(\tau)} = \left| \hat F^k_{{\rm PI}(\D_{cal})}(\tau) - \frac{1}{|\D_{cal}|}\sum_{\mathbf{x}_i \in \mathcal{X}_{cal}}(\hat{F}_k(\tau|\mathbf{x}_i))\right|
    \end{flalign*}%
\end{definition}

\item[CR-D-calibration] a competing-risks extension of D-calibration \citep{haider2020effective}: the normalized predicted incidence $\hat F_k(t_i\mid\mathbf{x}_i)/\hat F_k(\infty\mid\mathbf{x}_i)$ should be uniform on $[0,1]$ for subjects experiencing cause $k$. For each $\rho \in [0,1]$ the observed bin mass $\hat b^k_{[0,\rho]}$ (with a censoring correction) is compared to $\rho$, and the deviations are integrated over $\rho$ and summed over causes. It targets conditional calibration. Full definition in Def.~\ref{def:dcalestimation}.
\begin{definition}[CR $\hat{D}^{CR}_\alpha$-calibration]\label{def:dcalestimation}
    For a fixed $ \rho \in [0, 1]$, and event type $k$:
    \begin{multline*}
         \hat{b}^k_{[0, \rho]} = \frac{{\rm card}\left\{i ~\middle|~ \delta_i=k, \hat{F}_{k/\infty}(t_i\mid \mathbf{x}_i)\in [0, \rho]\right\}}{ \sum_{i}\{\hat F_k(\infty\mid \mathbf{x}_i)\}} \\
         + \frac{
            {\displaystyle\sum_{i}} \left\{\frac{\hat{F}_k(\infty\mid \mathbf{x}_i)\rho - \hat{F}_k(t_i\mid \mathbf{x}_i)}{\hat{S}(t_i\mid \mathbf{x}_i)} ~\middle|\!\begin{array}{l}
                 \small\delta_i=0  \\
                 \small\hat{F}_{k/\infty}(t_i\mid \mathbf{x}_i)\in[0,\rho])
            \end{array}\!\! \right\}
        }{
            {\sum_{i}}\{\hat F_k(\infty\mid \mathbf{x}_i)\}
        }
    \end{multline*}
    Where the first term counts individuals who experienced event $k$ and whose normalized prediction falls in $[0,\rho]$, scaled by the total predicted risk. The second term accounts for censored patients,  estimating their contribution.
    Then, we can define an estimator of the CR $D^{CR}_\alpha$-calibration as:
     \begin{flalign*}
        \hat{D}^{CR}_\alpha\text{-cal estimator} && \!\!\hat{D}^{CR}_\alpha \defeq \sum_{k=1}^K \! \left(\int_0^1\!\left| \hat{b}^k_{[0, \rho]} - \rho \right|^\alpha\! d\rho\right)^\frac{1}{\alpha}
\end{flalign*}
\end{definition}

\end{description}
For both, a value of $0$ is perfect calibration and lower is better; in the main text we report the gain over the marginal Aalen--Johansen estimator \citep{aalen_empirical_1978}.

\clearpage

\section{Details on the experimental methods}

\subsection{Competing Risks libraries}
Table \ref{tab:pythonlibraries} presents the libraries used to evaluate or download the datasets.

\begin{table*}[h!]
    \centering
    \caption{\textbf{Open-source libraries used in this benchmark.}}
    \label{tab:pythonlibraries}
    \begin{tabular}{ccc}
        \toprule
        Package & Language &used for \\
        \midrule
        pycox & Python &Downloading datasets, DeepHit Architecture \\
        lifelines & Python & Aalen--Johansen estimator \citep{aalen_empirical_1978} \\
        auton-survival & Python &Downloading Datasets \\
        hazardous & Python & Computing CR metrics (IBS, C-index, calibrations), SurvivalBoost \\
        cbioportal & R & Downloading Datasets \\
        cmprsks & R & Fine and Gray estimator \citep{fine_proportional_1999}, RSF \\
         &
    \end{tabular}
\end{table*}

\FloatBarrier

\subsection{Compute and Hyper parameter search}
\paragraph{Compute} To run the experiments, we had two setups, one for the CPU methods \citep{fine_proportional_1999, ishwaran_random_2014, alberge_survival_2025} and one for the GPUs ones.
The CPUs methods were ran onto 2x Intel Xeon E5-2660 v2 @ 2.20GHz with 40 cores and 256Go RAM. The GPUs ones were ran onto 2x Nvidia V100 32G, with the 2x AMD EPYC 7502 @ 2.50GHz processor, 128 cores, and 512G RAM.

\begin{center}
\small
\begin{longtable}{@{}p{0.25\linewidth}|l>{\raggedright\arraybackslash}p{0.44\linewidth}@{}}
    \caption{Hyper-parameter search space for every backbone and baseline. Each cell is searched with \texttt{RandomizedSearchCV} for 150 iterations over 3 stratified splits. \texttt{randint}/\texttt{loguniform}/\texttt{uniform} denote the \texttt{scipy.stats} distributions; a list denotes a categorical choice. The ranking weight $\alpha$ and temperature $\sigma$ are searched only for the HLRanking loss, which is the only loss that uses them.} \label{tab:hprealmlp}\\
        \toprule
        model & parameter & range \\
        \midrule
        \endfirsthead
        \toprule
        model & parameter & range \\
        \midrule
        \endhead
        \bottomrule
        \endfoot
        \bottomrule
        \endlastfoot
        \multicolumn{3}{c}{\textit{Backbones (each trained under all three losses)}} \\
        \midrule
        MLP & batch size & 128, 256 \\
         & dropout & None, 0.1, 0.2, 0.3 \\
         & epochs & \texttt{randint}(256, 1024) \\
         & lr & \texttt{loguniform}(1e-05, 0.001) \\
         & nb output times & 40, 60, 80, 100 \\
         & num nodes shared & [128, 128], [256, 256], [64, 64] \\
         & weight decay & \texttt{loguniform}(1e-05, 0.001) \\
         & alpha $^{\dagger}$ & \texttt{uniform}[0, 1] \\
         & sigma $^{\dagger}$ & \texttt{loguniform}(0.01, 10) \\
        \midrule
        RealMLP & base lr & \texttt{loguniform}(1e-05, 0.001) \\
         & batch size & 128, 256 \\
         & beta1 & \texttt{uniform}[0.1, 0.99] \\
         & beta2 & \texttt{uniform}[0.1, 0.99] \\
         & hidden sizes & [128, 128, 128], [256, 256, 256], [64, 64, 64], [128, 256, 128], [256, 128, 256], [64, 128, 64], [128, 64, 128], [256, 64, 256], [64, 256, 64] \\
         & n epochs & \texttt{randint}(256, 1024) \\
         & nb output times & 40, 60, 80, 100 \\
         & weight decay & \texttt{loguniform}(1e-05, 0.001) \\
         & alpha $^{\dagger}$ & \texttt{uniform}[0, 1] \\
         & sigma $^{\dagger}$ & \texttt{loguniform}(0.01, 10) \\
        \midrule
        TabM & batch size & 128, 256 \\
         & d block & 64, 128, 256 \\
         & epochs & \texttt{randint}(256, 1024) \\
         & k & 4, 8, 16 \\
         & lr & \texttt{loguniform}(1e-05, 0.001) \\
         & n blocks & 1, 2, 3, 4 \\
         & nb output times & 40, 60, 80, 100 \\
         & weight decay & \texttt{loguniform}(1e-05, 0.001) \\
         & alpha $^{\dagger}$ & \texttt{uniform}[0, 1] \\
         & sigma $^{\dagger}$ & \texttt{loguniform}(0.01, 10) \\
        \midrule
        SurvTRACE & attention probs dropout prob & \texttt{uniform}[0, 0.3] \\
         & batch size & 64, 128, 256 \\
         & patience & \texttt{randint}(3, 20) \\
         & hidden dropout prob & \texttt{uniform}[0, 0.3] \\
         & hidden size & 16, 32, 64, 128 \\
         & initializer range & \texttt{loguniform}(0.001, 0.1) \\
         & intermediate size & 64, 128, 256 \\
         & lr & \texttt{loguniform}(1e-05, 0.01) \\
         & max epochs & \texttt{randint}(256, 1024) \\
         & nb output times & 20, 40, 60, 80, 100 \\
         & num attention heads & 1, 2, 4, 8 \\
         & num hidden layers & 1, 2, 3, 4 \\
         & betas & (0.9, 0.999), (0.9, 0.95), (0.95, 0.999), (0.8, 0.999) \\
         & weight decay & \texttt{loguniform}(1e-06, 0.01) \\
         & alpha $^{\dagger}$ & \texttt{uniform}[0, 1] \\
         & sigma $^{\dagger}$ & \texttt{loguniform}(0.01, 10) \\
        \midrule
        \multicolumn{3}{c}{\textit{Baselines (each trained with its own fixed loss)}} \\
        \midrule
        Aalen--Johansen \citep{aalen_empirical_1978} & \multicolumn{2}{c}{no hyperparameters (population-level estimator)} \\
        \midrule
        Fine and Gray \citep{fine_proportional_1999} & max iter & \texttt{randint}(10, 100) \\
         & max samples & 50000 \\
         & n components & 5, 10, 20, 30, 50 \\
         & tol & \texttt{loguniform}(1e-08, 0.0001) \\
        \midrule
        Random Survival Forests & max depth & None, 4, 8, 12, 16, 24 \\
         & max features & sqrt, log2, 0.3, 0.5, 0.8, None \\
         & min samples leaf & \texttt{randint}(1, 25) \\
         & min samples split & \texttt{randint}(2, 30) \\
         & n estimators & \texttt{randint}(100, 1000) \\
        \midrule
        SurvivalBoost & hard zero fraction & \texttt{uniform}[0, 0.5] \\
         & learning rate & \texttt{loguniform}(0.01, 0.3) \\
         & max depth & \texttt{randint}(2, 10) \\
         & max leaf nodes & 15, 31, 63, 127 \\
         & min samples leaf & \texttt{randint}(5, 100) \\
         & n horizons per observation & \texttt{randint}(1, 5) \\
         & n iter & \texttt{randint}(50, 200) \\
         & n iter before feedback & \texttt{randint}(5, 50) \\
         & n time grid steps & 50, 100, 200 \\
        \midrule
        SurvTRACE  & attention probs dropout prob & \texttt{uniform}[0, 0.3] \\
         & batch size & 64, 128, 256 \\
         & patience & \texttt{randint}(3, 20) \\
         & hidden dropout prob & \texttt{uniform}[0, 0.3] \\
         & hidden size & 16, 32, 64, 128 \\
         & initializer range & \texttt{loguniform}(0.001, 0.1) \\
         & intermediate size & 64, 128, 256 \\
         & lr & \texttt{loguniform}(1e-05, 0.01) \\
         & max epochs & \texttt{randint}(50, 500) \\
         & num attention heads & 1, 2, 4, 8 \\
         & num hidden layers & 1, 2, 3, 4 \\
         & betas & (0.9, 0.999), (0.9, 0.95), (0.95, 0.999), (0.8, 0.999) \\
         & weight decay & \texttt{loguniform}(1e-06, 0.01) \\
         & quantile horizons & evenly spaced quantiles, $n \in \{3, 10, 20, 40, 60\}$ \\
        \midrule
        Deep Survival Machines & batch size & 128, 256 \\
         & discount & 0.5, 0.75, 1.0 \\
         & distribution & Weibull, LogNormal \\
         & epochs & \texttt{randint}(50, 300) \\
         & k & 3, 4, 6, 8 \\
         & layers & None, [64], [128], [64, 64], [128, 128], [256, 256] \\
         & lr & \texttt{loguniform}(0.0001, 0.01) \\
        \midrule
        DeSurv & batch size & 128, 256 \\
         & epochs & \texttt{randint}(100, 500) \\
         & hidden dim & 32, 64, 128, 256 \\
         & lr & \texttt{loguniform}(0.0001, 0.01) \\
        \midrule
        Neural Fine-Gray & batch size & 128, 256, 512 \\
         & dropout & \texttt{uniform}[0, 0.5] \\
         & epochs & \texttt{randint}(100, 1000) \\
         & layers & (50, 50), (100, 100), (100, 100, 100), (200, 200), (50, 100, 50), (200, 100, 200) \\
         & layers surv & (50,), (100,), (100, 100), (200,) \\
         & lr & \texttt{loguniform}(0.0001, 0.01) \\
         & patience max & \texttt{randint}(5, 30) \\
         & weight decay & \texttt{loguniform}(1e-05, 0.01) \\
        \midrule
        DQS & batch size & 128, 256 \\
         & epochs & \texttt{randint}(100, 500) \\
         & hidden dim & 64, 128, 256 \\
         & lr & \texttt{loguniform}(0.0001, 0.01) \\
         & n bins & 20, 50, 100 \\
        \midrule
\end{longtable}
\end{center}

\paragraph{Selected ranking weight of HLRanking}
Table~\ref{tab:alphaselected} reports the weight $\alpha$ of $\mathcal{L}_{HLR} = \alpha\,\mathcal{L}_{HL} +
(1-\alpha)\,\mathcal{L}_{R}$ (Sec.~\ref{sec:crlosses}) selected by the search above
(\texttt{uniform}$[0, 1]$) for HLRanking, for each backbone and for both selection criteria.
\emph{Conventions.} DeepHit \citep{lee_deephit_2018} writes the loss as $\mathcal{L}_{HL} +
\alpha_{\mathrm{DH}}\,\mathcal{L}_{R}$ with $\alpha_{\mathrm{DH}} \geq 0$ on the ranking term. We use the
pycox implementation \citep{kvamme2019pycox} in every backbone, which instead takes the convex
combination above: $\alpha$ weights the likelihood, $\alpha = 1$ is pure likelihood and $\alpha = 0$ pure
ranking. Up to an overall scale absorbed by the learning rate, the two are related by
$\alpha_{\mathrm{DH}} = (1-\alpha)/\alpha$, so a \emph{small} $\alpha$ here means a \emph{large} ranking
weight.
\emph{What the selected values imply.} Tuned on IBS, the median $\alpha$ is $0.59$ over all backbones
($\alpha_{\mathrm{DH}} \approx 0.7$): the likelihood carries more weight than the ranking term, which is
down-weighted but not switched off. Tuned on the C-index, the median drops to $0.29$
($\alpha_{\mathrm{DH}} \approx 2.4$) and the ranking term dominates, on RealMLP almost entirely (median
$\alpha \approx 0.003$). Each criterion thus pulls the ranking weight in the direction that favours
it, which is the trade-off Tab.~\ref{tab:hlrankingselection} measures.

\begin{table}[h!]
    \centering\small
    \caption{\textbf{Ranking weight $\alpha$ selected for HLRanking}, by selection criterion and
    backbone: median, quartiles and range.}
    \label{tab:alphaselected}
    \begin{filecontents*}{figures/alpha_quantiles.csv}
tuning,architecture,min,q25,median,q75,max
ibs,MLP,0.19705428018563964,0.5627831242980726,0.5627831242980726,0.8217185635803377,0.9509526110553941
ibs,RealMLP,0.09345945644411735,0.5216074005622916,0.657019774194,0.821499744565374,0.983426242260642
ibs,TabM,0.03999280190071697,0.20486576218044572,0.5777094243168404,0.7316337132869178,0.9040443929009577
ibs,SurvTRACE,0.15204702504442202,0.46052405679998054,0.6565344873343768,0.7442759301031997,0.9854914276432976
ibs,All architectures,0.03999280190071697,0.4227346139820079,0.5910653808339902,0.7902396708582071,0.9854914276432976
c_index,MLP,0.030452964941600325,0.19942524496232983,0.45651989126265535,0.545656139468861,0.8286569145557378
c_index,RealMLP,0.0006642185902044906,0.0012075717368487116,0.003198599496646859,0.6596123999640857,0.8880437054177104
c_index,TabM,0.009240073486540767,0.025662718054531575,0.20486576218044572,0.6747523222590207,0.9979622513286734
c_index,SurvTRACE,0.06099500678157388,0.2153982043432382,0.292177365167522,0.519642829449547,0.9686880692761627
c_index,All architectures,0.0006642185902044906,0.030452964941600325,0.292177365167522,0.6330620068193471,0.9979622513286734
\end{filecontents*}
\pgfplotstabletypeset[col sep=comma,
        columns={tuning,architecture,median,q25,q75,min,max},
        every head row/.style={before row=\toprule, after row=\midrule},
        every last row/.style={after row=\bottomrule},
        every row no 4/.style={before row=\cmidrule(l){2-7}},
        every row no 5/.style={before row=\midrule},
        every row no 9/.style={before row=\cmidrule(l){2-7}},
        fixed, zerofill, precision=2,
        columns/tuning/.style={column name=Selection, string type, column type=l,
            assign cell content/.code={%
                \ifnum\pgfplotstablerow=0 \pgfkeyssetvalue{/pgfplots/table/@cell content}{IBS}%
                \else\ifnum\pgfplotstablerow=5 \pgfkeyssetvalue{/pgfplots/table/@cell content}{C-index}%
                \else\pgfkeyssetvalue{/pgfplots/table/@cell content}{}\fi\fi}},
        columns/architecture/.style={column name=Backbone, string type, column type=l,
            string replace={All architectures}{All backbones}},
        columns/median/.style={column name=Median},
        columns/q25/.style={column name=$Q_1$},
        columns/q75/.style={column name=$Q_3$},
        columns/min/.style={column name=Min},
        columns/max/.style={column name=Max},
    ]{figures/alpha_quantiles.csv}
\end{table}

\subsection{Training-time IPCW sensitivity: Kaplan-Meier vs. Cox vs. stratified}\label{sec:ipcwsensitivity}
To justify the marginal Kaplan-Meier default used for the CRNLL loss's training-time IPCW (Sec.~\ref{sec:crlosses}), we retrained CRNLL under three training-time censoring-weighting schemes -- marginal Kaplan-Meier \citep{kaplan_nonparametric_1958}, Cox \citep{cox1972regression}, and a covariate-stratified Kaplan-Meier (a piecewise-constant-in-covariates middle ground) -- on all four backbones, holding the evaluation-side IPCW fixed at marginal Kaplan-Meier throughout so that only the training-time weighting changes.

Cox is markedly less reliable than the other two schemes. Where Kaplan-Meier scores all 94 (dataset, backbone) cells and the stratified estimator scores 93/94, Cox produces a usable fit on only 35/94 (63\% missing, due to non-convergence). On the 35 cells where all three schemes completed, splitting at a 20\% relative IBS gain over Kaplan-Meier gives a sharp split: 24 cells are stable (median gain $+0.0\%$ for Cox, $+0.4\%$ for stratified), while on the remaining 11 cells Cox's IBS is on average $+139.4\%$ worse than Kaplan-Meier's -- and in every one of these 11 cells Cox is worse, never better -- while the stratified estimator stays within a couple of percent of Kaplan-Meier even on this same subset (Table~\ref{tab:ipcwsensitivity}). Cox is also slower to fit: a median $1.75\times$ and up to $6.7\times$ the fitting time of Kaplan-Meier, on the cells where both complete. We found no consistent relationship between a dataset's censoring rate and whether Cox is stable on it: the blown-up cells span censoring rates from 35\% to 98\%, and some of the most heavily censored datasets in the benchmark (e.g. nwtco, smarto) are among the most stable for Cox.

\begin{table}[h!]
    \centering
    \caption{\textbf{A Cox censoring model is unreliable here, while a covariate-stratified one
    reproduces the marginal estimate.} CRNLL training-time IPCW, Cox against Kaplan-Meier against
    stratified Kaplan-Meier, on the 35 (dataset, backbone) cells where all three schemes produced a
    usable fit, split at $|$IBS gain over Kaplan-Meier$| = 20\%$.}
    \label{tab:ipcwsensitivity}
    \begin{tabular}{lccc}
        \toprule
        & $n$ cells & median Cox vs.\ KM & median stratified vs.\ KM \\
        \midrule
        Stable & 24 & $+0.0\%$ & $+0.4\%$ \\
        Blown up ($>20\%$ worse) & 11 & $+139.4\%$ & $+0.2\%$ \\
        \bottomrule
    \end{tabular}
\end{table}

We read this as supporting the marginal-Kaplan-Meier default used throughout the paper: a stable, covariate-adjusted alternative (stratified Kaplan-Meier) reproduces it almost everywhere it is estimable, while Cox's instability under this benchmark is a property of that estimator on these datasets, not evidence that a covariate-conditional censoring model would change our conclusions if it were reliably estimable.

\section{Detailed Experimental Results}
\subsection{Computing effect sizes: How much the loss matters, and where}\label{sec:effectsizes}

Section \ref{sec:results} compares the importance of the two axes loss/backbone with three numbers, obtained from the per-(dataset, backbone,
loss) mean IBS gains that also produce Fig.~\ref{fig:ibsgain}. Restricted to the 12 cells of the
loss $\times$ backbone matrix, so that both axes are free, and to the 20 tabular datasets of
Table~\ref{tab:datasetsconsidered}. The \emph{loss effect} is, for each (backbone, dataset) pair, the
spread over the three losses, $\max - \min$; the \emph{backbone effect} is, for each (loss, dataset)
pair, the spread over the four backbones. Both are in IBS-gain units and are paired within a dataset,
which is what makes them comparable. Coverage is 229 of the 240 cells, and a spread is taken only where
every member of its group is present, so the two effects rest on $n=75$ and $n=49$ groups respectively.
Table \ref{tab:effectsizes} details the corresponding estimates.

\begin{table}[h!]
    \centering\small
    \caption{\textbf{The loss separates the models twice as much as the backbone on competing-risks
    data, and slightly less than the backbone on single-event data.} Median spread in IBS gain, with
    the interquartile range across groups. Pooling the two families averages two opposite regimes and
    hides the effect, which is why the main text reports them apart.}
    \label{tab:effectsizes}
    \begin{tabular}{lrrrrr}
\toprule
Datasets & \multicolumn{2}{c}{Loss effect} & \multicolumn{2}{c}{Backbone effect} & Ratio \\
\cmidrule(r){2-3}\cmidrule(r){4-5}\cmidrule(l){6-6}
 & median & IQR & median & IQR & \\
\midrule
Competing risks & 0.053 & [0.020, 0.084] & 0.027 & [0.021, 0.062] & 1.93 \\
Single event & 0.072 & [0.045, 0.132] & 0.086 & [0.039, 0.208] & 0.83 \\
All 20 & 0.065 & [0.030, 0.100] & 0.059 & [0.023, 0.136] & 1.10 \\
\bottomrule
\end{tabular}

\end{table}

\begin{table}[h!]
    \centering\footnotesize
    \caption{\textbf{Per-dataset view behind Tab.~\ref{tab:effectsizes}.} For each dataset, the median
    spread over the three losses and over the four backbones, and which loss attains the best median
    gain. The loss spread exceeds the backbone spread on most competing-risks datasets and on fewer of
    the single-event ones.}
    \label{tab:perdatasetspreads}
    \begin{tabular}{llrrl}
\toprule
Dataset & Type & Loss spread & Backbone spread & Best loss \\
\midrule
framingham & CR & 0.036 & 0.059 & CRNLL \\
metabric\_competing & CR & 0.068 & 0.022 & CRNLL \\
mimic & CR & 0.012 & -- & CRNLL \\
mimiciii & CR & 0.011 & 0.020 & CRNLL \\
seer & CR & 0.043 & 0.021 & CRNLL \\
seer\_digestion & CR & 0.029 & -- & CRNLL \\
pbc & CR & 0.100 & 0.060 & CRNLL \\
loan & CR & 0.038 & 0.033 & CRNLL \\
eicu & CR & 0.298 & 0.213 & CRNLL \\
tcga & CR & 0.103 & 0.079 & CRNLL \\
\midrule
flchain & single & 0.067 & 0.197 & CRNLL \\
hdfail & single & 0.110 & 0.146 & CRNLL \\
gbsg & single & 0.126 & 0.218 & CRNLL \\
metabric & single & 0.052 & 0.028 & CRNLL \\
msk\_2021 & single & 0.066 & 0.055 & CRNLL \\
msk\_2026 & single & 0.064 & 0.052 & CRNLL \\
nwtco & single & 0.155 & 3.136 & CRNLL \\
smarto & single & 0.137 & 0.256 & CRNLL \\
support & single & 0.078 & 0.120 & CRNLL \\
challenge\_ens & single & 0.041 & 0.067 & CRNLL \\
\bottomrule
\end{tabular}

\end{table}

\FloatBarrier

\subsection{Comparing the different model-selection metrics}\label{sec:allmetrics}

Here we give results for the six model-selection metrics of Sec.~\ref{sec:crmetrics} on the tabular data.

\begin{figure}[h!]
    \includegraphics[width=\linewidth]{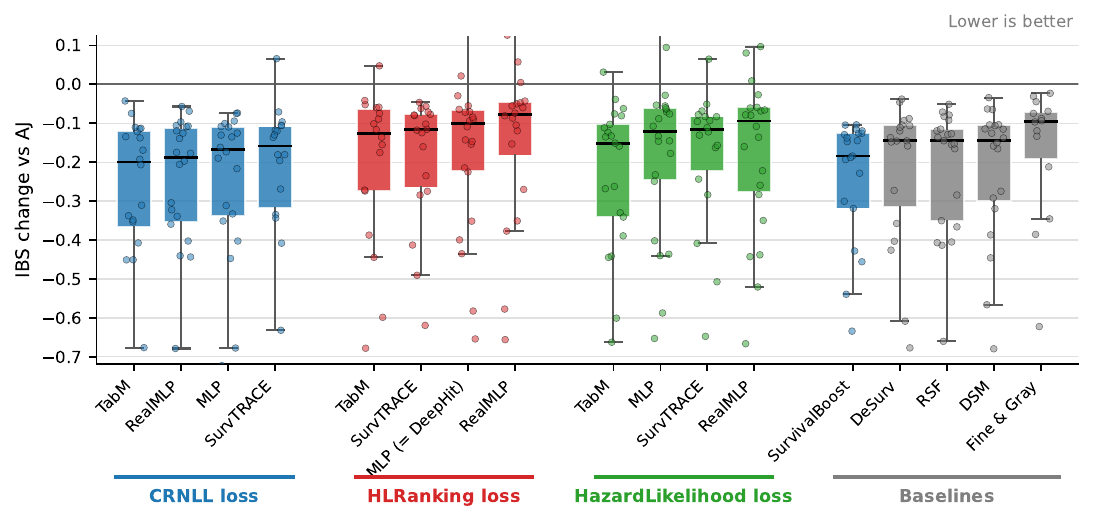}
    \caption{\textbf{IBS results.} (reproducing Fig.~\ref{fig:ibsgain} of the main text for ease of read) IBS gain over the Aalen--Johansen estimator on all 20 tabular datasets.}
\end{figure}

\begin{figure}[h!]
    \includegraphics[width=\linewidth]{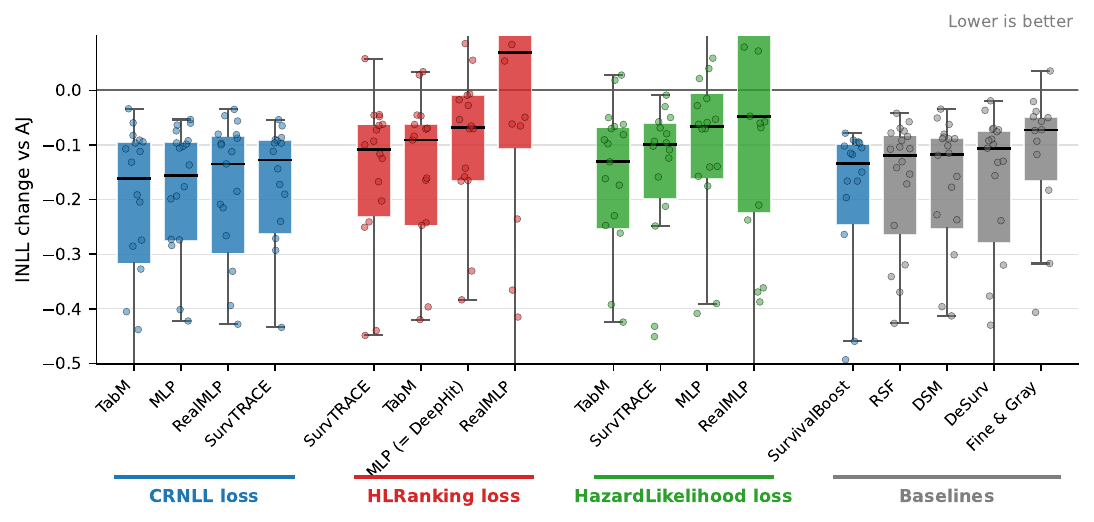}
    \caption{\textbf{The second proper scoring rule orders the losses as the Brier score does.}
    Integrated negative log-likelihood against the Aalen--Johansen estimator.
    CRNLL is lowest on every backbone.}
    \label{fig:inllgain}
\end{figure}

\begin{figure}[h!]
    \includegraphics[width=\linewidth]{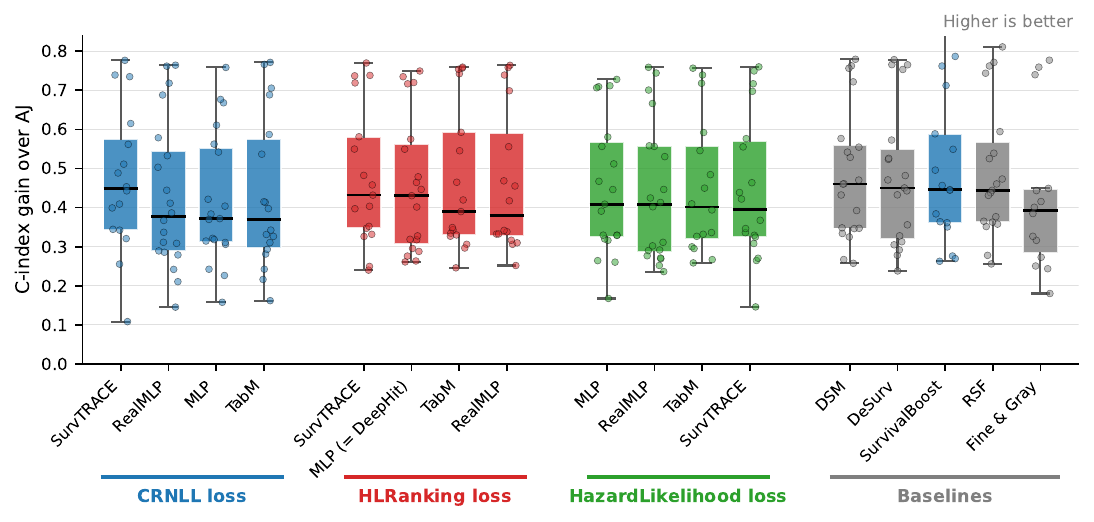}
    \caption{\textbf{The C-index does not separate well the models.} Gain in C-index over the
    Aalen--Johansen estimator, higher is better. The boxes overlap across losses, backbones and baselines alike.}
    \label{fig:cindexgain}
\end{figure}

\begin{figure}[h!]
    \includegraphics[width=\linewidth]{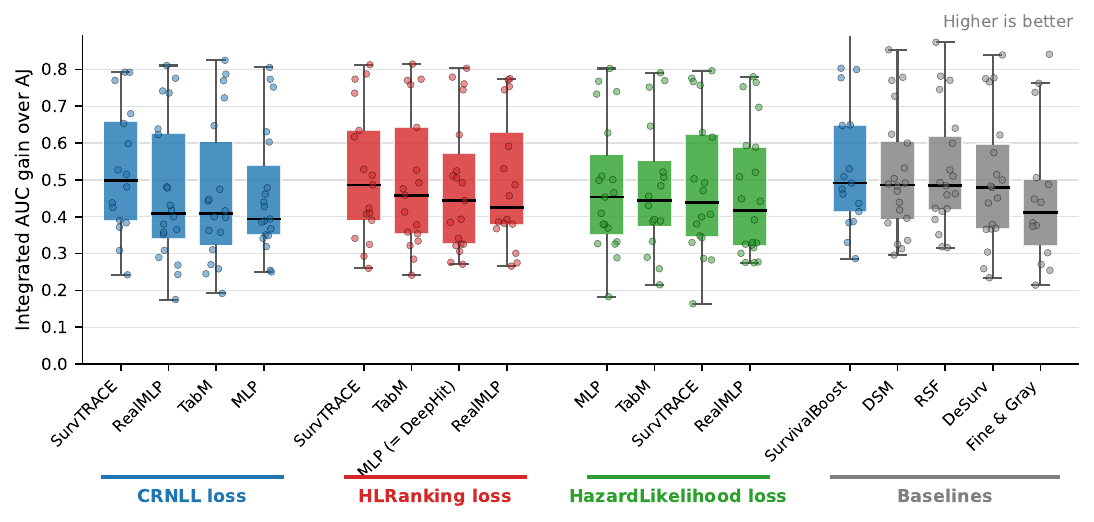}
    \caption{\textbf{Integrated AUC also gives a muddied picture.} Gain in
    integrated time-dependent AUC over the Aalen--Johansen estimator, higher is better. As for the C-index, models are not well separated.}
    \label{fig:aucgain}
\end{figure}

The six metrics fall into two groups: strictly proper scoring rules and others. Results of whether they efficiently tell the models apart are related to their properness. The two proper scoring rules agree: the Brier score
(Fig.~\ref{fig:ibsgain}) and the integrated negative log-likelihood (Fig.~\ref{fig:inllgain}) both put
the CRNLL group ahead of the hazard-based ones on every backbone. The two ranking metrics behave similarly but do not distinguish models or losses: neither the C-index (Fig.~\ref{fig:cindexgain}) nor the integrated AUC (Fig.~\ref{fig:aucgain})
separates any model from any other, the boxes overlapping almost completely across all twelve cells and
the nine baselines.
Note that AUC is proper (it is minimized by the true generative mechanism), but not strictly proper (the true generative mechanism is not the \textsl{only} minimizer. Experimental results show that its integration over time is gives more muddied signal that than a strictly proper scoring rule (IBS or INLL). We discuss calibration metrics below.

\FloatBarrier

\paragraph{Calibration}
We detail below the two calibration criteria. While calibration is an important topic, it comes with different definitions in the case of survival analysis (whether single-event or competing risks), which are difficult to estimate:
\begin{description}

\item[AJ-calibration]
The AJ-calibration (Definition~\ref{def:ajcal}) measures the area between a
model's marginal predictions (i.e., the mean predicted cumulative incidence
across samples) and the Aalen-Johansen estimator; a model is considered perfectly
calibrated when this quantity is near zero~\citep{alberge2026calibrationsurvivalmodelscompeting}.
The boxplots confirm that models trained with the CRNLL loss are the best calibrated overall, while
those trained with the HLRanking loss exhibit the largest calibration errors.

\item[D-calibration]
We additionally report the D-calibration gain over the Aalen-Johansen estimator~\citep{aalen_empirical_1978}. Despite the large variance across datasets, a tendency emerges: proper scoring rule (PSR) based losses, and the HazardLikelihood loss in particular, yield better-calibrated models. As with AJ-calibration, a smaller D-calibration value indicates better calibration~\citep{alberge2026calibrationsurvivalmodelscompeting}.

\end{description}

\begin{figure}[h!]
    \includegraphics[width=\linewidth]{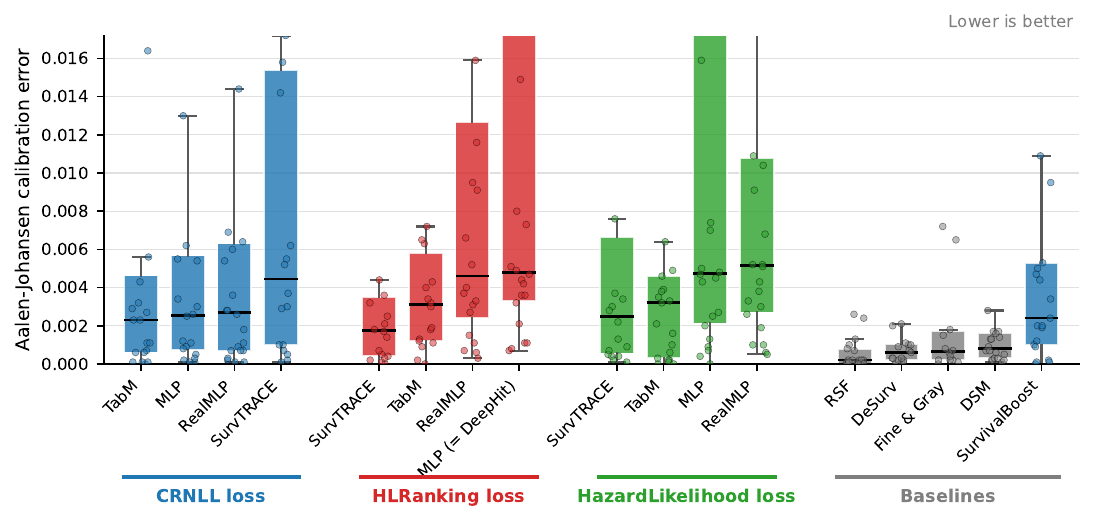}
    \caption{\textbf{On raw marginal calibration too, CRNLL is the best calibrated and HLRanking the
    worst.} Area between the Aalen--Johansen estimator \citep{aalen_empirical_1978} and one model.}
    \label{fig:aj_calib_raw}\label{fig:aj_calib_gain}
    \smallskip

    \includegraphics[width=\linewidth]{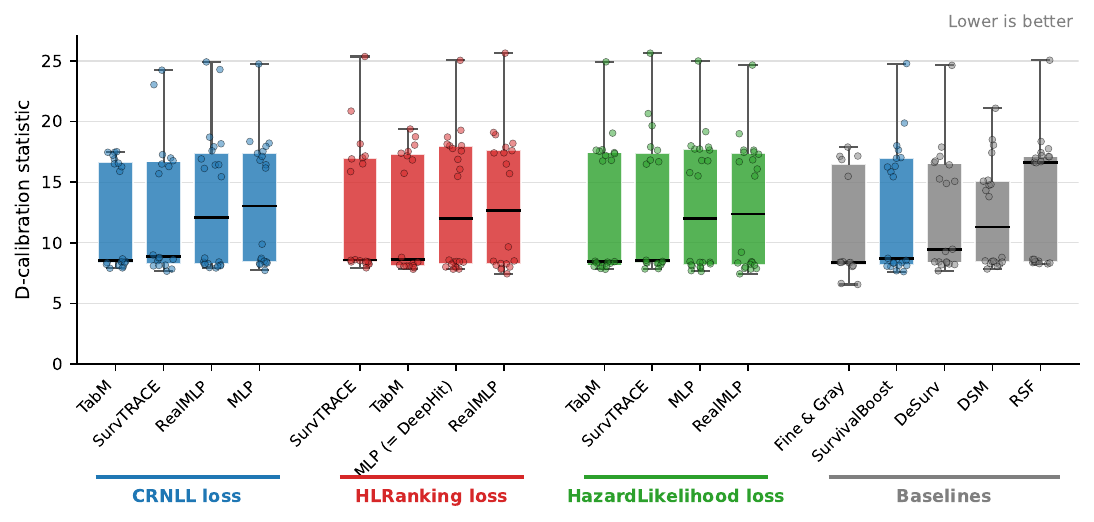}
    \caption{\textbf{On D-calibration the three losses overlap heavily, the spread across
    datasets being far larger than any difference between them.} Raw D-calibration statistic}
    \label{fig:dcalib}
\end{figure}

The intention behind the use of calibration metrics is typically to judge how far a model is from good probabilities. In terms of pure model selection (comparing multiple models), our results suggest that strictly proper scoring rules should probably be preferred.

\paragraph{Is the C-index flatter on competing risks than on single-event data?}
We tested whether the C-index separates models less on the 10 competing-risks datasets than on
the 10 single-event ones, reusing the spreads of Sec.~\ref{sec:effectsizes} on the C-index gain at
the median time (Tab.~\ref{tab:cindexcrsingle}). It does not. The spread across losses, across
backbones, or across all cells of a dataset is about the same in both families, and a one-sided
Mann--Whitney test of a smaller spread on competing risks is far from significant ($p \geq 0.43$).
Relative to the IBS spread on the same group, the C-index even moves more on competing risks,
mostly because the IBS itself separates the models less there. Nor does the C-index recover the
IBS ordering in either family: the per-dataset Spearman correlation between the two rankings of the
cells has median $-0.23$ on competing risks and $-0.06$ on single-event data ($p = 0.52$). With 7 to
10 datasets per family these tests have little power, but they give no sign that the C-index is
less informative in one family: it is equally flat in both.

\begin{table}[h!]
    \centering\footnotesize
    \caption{\textbf{The C-index separates models about as little on competing risks (CR) as on
    single-event data.} Median spread in C-index gain at the median time, with interquartile range,
    over the groups of Sec.~\ref{sec:effectsizes}; $p$: one-sided Mann--Whitney test of a smaller
    spread on CR. Right: the same spread divided by the IBS spread of the same group.}
    \label{tab:cindexcrsingle}
    \begin{tabular}{lccccc}
\toprule
 & \multicolumn{3}{c}{C-index spread} & \multicolumn{2}{c}{C-index / IBS spread} \\
\cmidrule(lr){2-4}\cmidrule(lr){5-6}
Spread across & CR & single event & $p$ & CR & single event \\
\midrule
the 3 losses & $0.032$ \scriptsize{[0.015, 0.067]} & $0.035$ \scriptsize{[0.015, 0.068]} & $0.43$ & $0.63$ \scriptsize{[0.46, 1.25]} & $0.50$ \scriptsize{[0.16, 1.05]} \\
the 4 backbones & $0.054$ \scriptsize{[0.026, 0.113]} & $0.038$ \scriptsize{[0.024, 0.075]} & $0.69$ & $1.09$ \scriptsize{[0.76, 4.09]} & $0.44$ \scriptsize{[0.14, 1.40]} \\
all cells of a dataset & $0.113$ \scriptsize{[0.061, 0.182]} & $0.082$ \scriptsize{[0.048, 0.133]} & $0.67$ & $1.11$ \scriptsize{[0.62, 1.76]} & $0.60$ \scriptsize{[0.18, 0.74]} \\
\bottomrule
\end{tabular}

\end{table}

\FloatBarrier

\subsection{Dataset stratification: multiple events or not, amount of censoring}

Here we study how results vary along different dataset characteristics, namely competing risks versus single-event survival analysis, as well as amount of censoring.

\begin{figure}[h!]
    \centering
    \includegraphics[width=\linewidth]{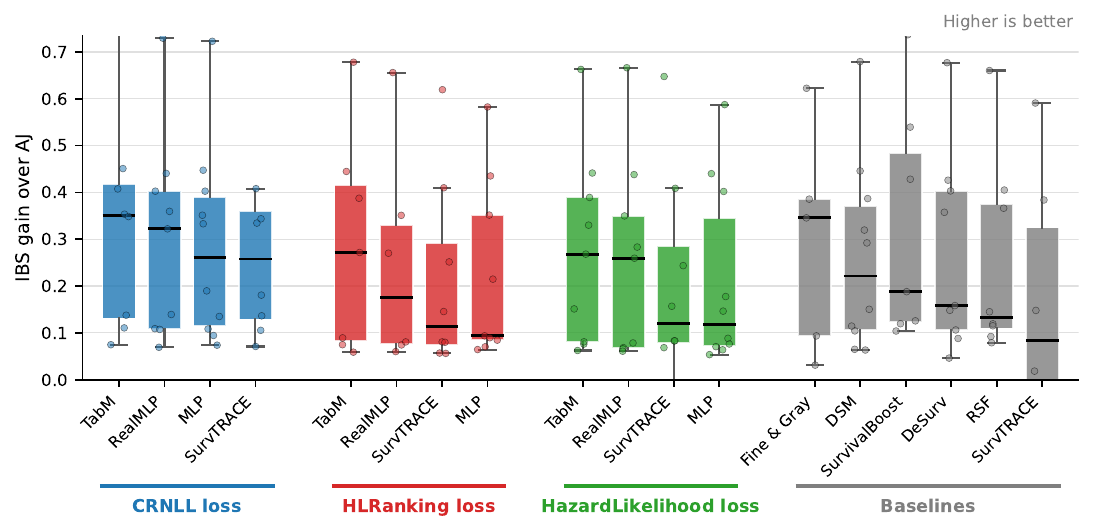}
    \caption{\textbf{The ordering is unchanged on the competing-risks datasets alone.} IBS gain over
    the Aalen--Johansen estimator, restricted to the 10 competing-risks datasets. Layout as in
    Fig.~\ref{fig:ibsgain}.}
    \label{fig:ibsgaincronly}
\end{figure}

\paragraph{Multiple events or not}
Figure \ref{fig:ibsgaincronly} gives the per-dataset IBS gain over the Aalen--Johansen estimator \citep{aalen_empirical_1978}
for each competing risks dataset. Results folow the trend observed in
Fig.~\ref{fig:ibsgain}, but the effect is even more marked in pure competing-risk settings: the CRNLL loss consistently achieves the largest gains across datasets and architectures and the HLRanking that struggles the most.

\begin{table}[h!]
    \centering\footnotesize
    \caption{\textbf{The advantage of the cumulative-incidence loss is far wider on competing-risks
    data.} Median IBS gain over Aalen--Johansen for each cell of the matrix, best backbone per row in
    bold. On competing risks CRNLL leads the nearest hazard-based loss by a large margin on every
    backbone; on single-event data the three losses fall much closer together.}
    \label{tab:lossbackbonemedians}
    \begin{tabular}{lrrrrrrrr}
\toprule
 & \multicolumn{4}{c}{Competing risks (10 datasets)} & \multicolumn{4}{c}{Single event (10 datasets)} \\
\cmidrule(r){2-5}\cmidrule(l){6-9}
Loss & MLP & RealMLP & TabM & SurvTRACE & MLP & RealMLP & TabM & SurvTRACE \\
\midrule
CRNLL & 0.261 & 0.250 & \textbf{0.351} & 0.258 & 0.146 & 0.150 & \textbf{0.152} & 0.125 \\
HazardLikelihood & 0.117 & 0.198 & \textbf{0.268} & 0.120 & 0.084 & 0.043 & \textbf{0.128} & 0.117 \\
HLRanking & 0.120 & 0.093 & \textbf{0.180} & 0.120 & 0.063 & 0.049 & \textbf{0.126} & 0.117 \\
\bottomrule
\end{tabular}

\end{table}

\begin{figure}[h!]
    \centering
    \includegraphics[width=0.85\linewidth]{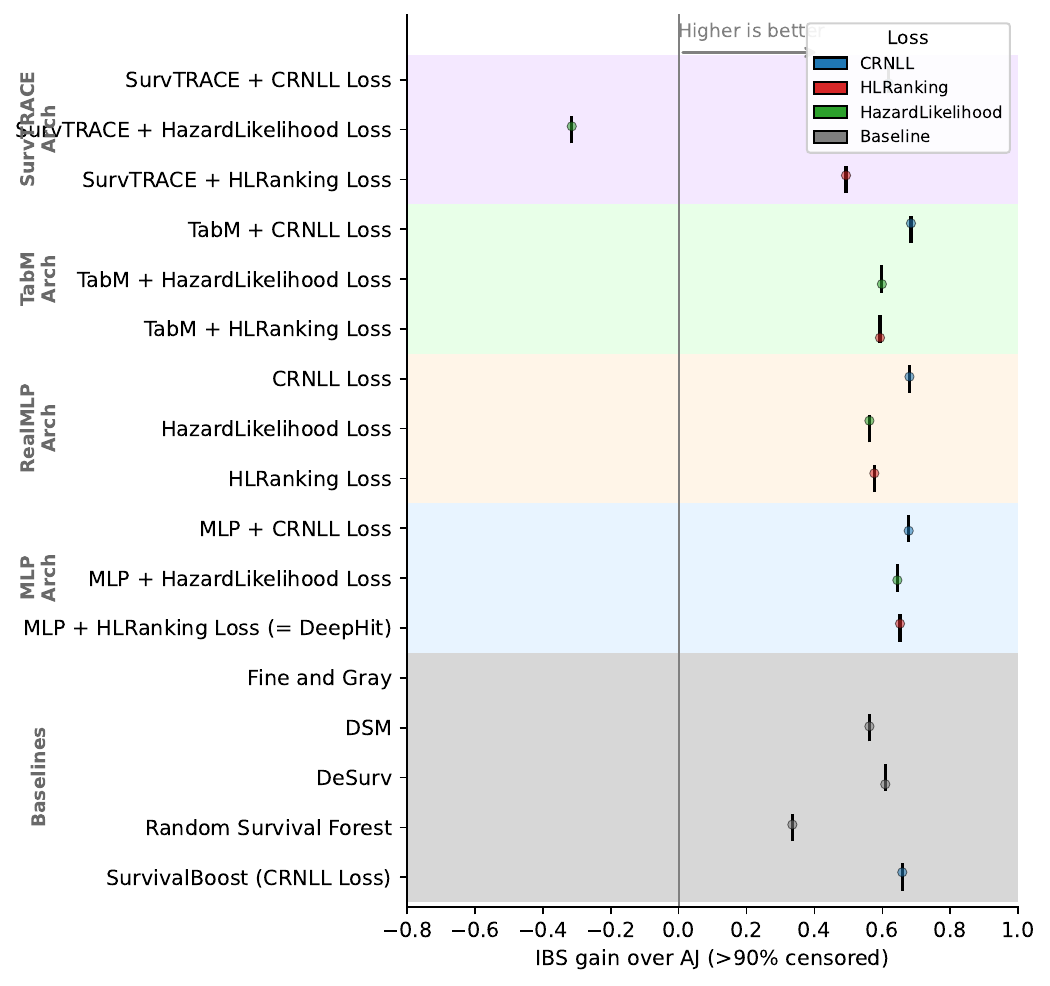}
    \caption{\textbf{Under heavy censoring the advantage of the CRNLL loss persists and widens.}
    IBS gain over the Aalen--Johansen estimator on the datasets censored above $90\%$, grouped by
    backbone so that the three losses can be compared within each block; higher is better. CRNLL
    is ahead of both hazard-based losses on every backbone, while those two remain close to each
    other, as in Fig.~\ref{fig:ibsgain}. Two contrasts with the full collection are worth noting:
    the random survival forest, which is competitive overall, falls well behind the neural models
    trained with CRNLL here, and one cell (SurvTRACE with the HazardLikelihood loss) degrades to worse
    than the Aalen--Johansen reference.}
    \label{fig:ibsgainhighcens}
\end{figure}

\paragraph{Amount of censoring}
Heavy censoring is the regime in which competing-risks models are hardest to fit and in which the
choice of a loss adapted to the censoring in competing risk is most important, so we also isolate the most heavily censored
datasets of the collection (Fig.~\ref{fig:ibsgainhighcens}). The figure shows a widening of the benefits observed for CRNLL on the full dataset.

\FloatBarrier

\subsection{C-indexes at different time points}\label{sec:cindexes}
\begin{figure}[tp!]
\begin{subfigure}{\textwidth}
    \includegraphics[width=\linewidth]{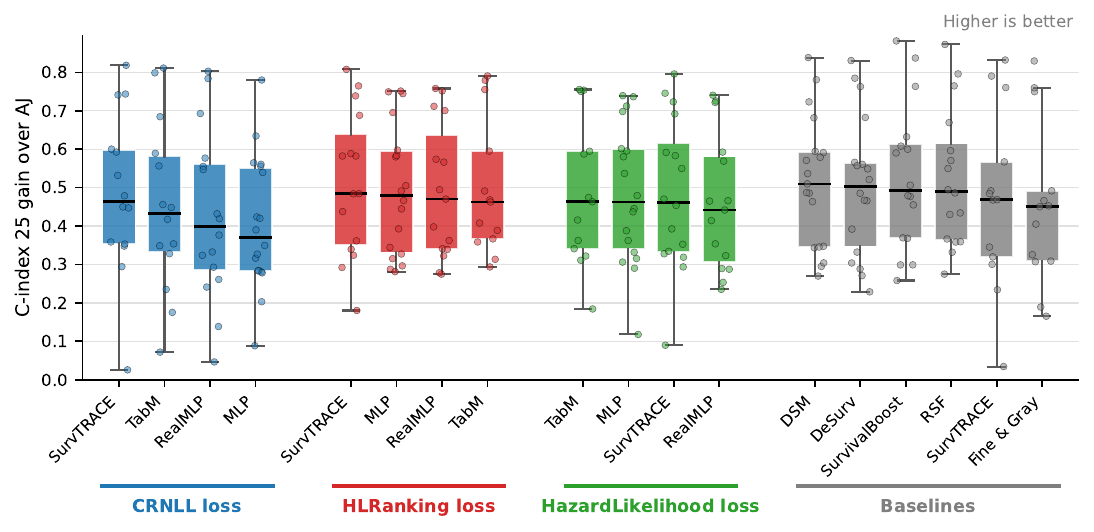}\vspace*{-1ex}
    \caption{\textbf{t = 25th percentile.}}\label{fig:cindex25gain}
\end{subfigure}
\begin{subfigure}{\textwidth}
    \includegraphics[width=\linewidth]{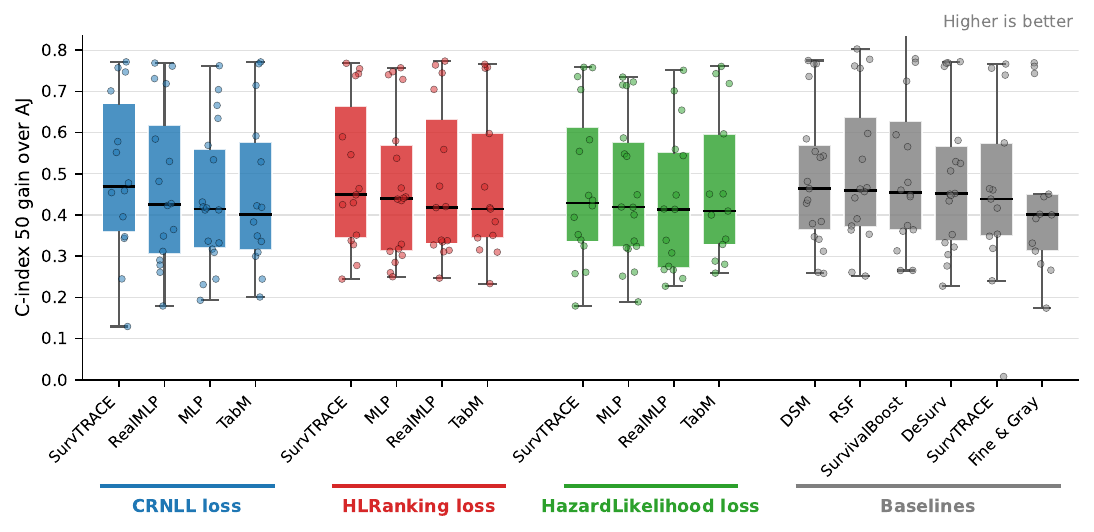}\vspace*{-1ex}
    \caption{\textbf{t = 50th percentile.}}\label{fig:cindex50gain}
\end{subfigure}
\begin{subfigure}{\textwidth}
    \includegraphics[width=\linewidth]{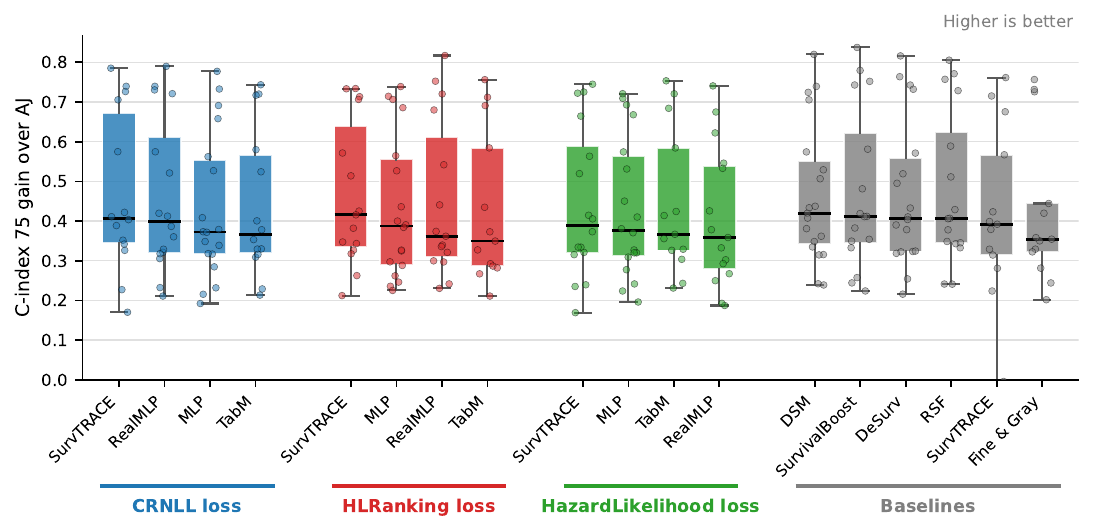}\vspace*{-1ex}
    \caption{\textbf{t = 75th percentile.}}\label{fig:cindex75gain}
\end{subfigure}
    \caption{\textbf{C-index at given times separates neither the loss nor the backbone}
    Gain in C-index over the Aalen--Johansen estimator for different time horizons. Layout as in
    Fig.~\ref{fig:ibsgain}.}\label{fig:cindexquantiles}
\end{figure}

The C-index is a ranking metric evaluated at a specific time point. The main
text summarises the 25th-quantile result; every C-index figure lives here. We show
the gain over the Aalen-Johansen estimator for all three quantiles (25th, 50th, 75th).
While a marginal advantage of the HLRanking loss is visible at the 25th quantile
(Fig.~\ref{fig:cindex25gain}), no clear trend emerges at the 50th or 75th
quantiles (Figs.~\ref{fig:cindex50gain},~\ref{fig:cindex75gain}). Here we see that even restricting C-index at given time points does not lead it to bring light on relative benefits of either
architectures or loss functions.

\subsection{HLRanking: tuning on its own target metric}\label{sec:hlrankingtuning}
The HLRanking results in the main text select hyperparameters on validation IBS, which leaves open the possibility that HLRanking's ranking term is simply undertuned for the metric it is designed to optimize. Table~\ref{tab:hlrankingselection} reports, for each of the four backbones, HLRanking's own IBS and C-index (50th quantile) under the two selection criteria of Sec.~\ref{sec:crlosses} (IBS-selected vs.\ C-index-selected), paired at the (dataset, backbone) level.

\begin{table}[h!]
    \centering
    \caption{\textbf{Tuning HLRanking on the C-index buys almost no C-index and costs a great deal
    of IBS.} IBS-selected against C-index-selected hyperparameters, paired per (dataset, backbone),
    with $\alpha$ and $\sigma$ searched jointly (Sec.~\ref{sec:crlosses}). $n$ = number of datasets with both selections
    available for that backbone. Median $\Delta$ = C-index-selected minus IBS-selected.}
    \label{tab:hlrankingselection}
    \begin{tabular}{lccccc}
        \toprule
        Backbone & $n$ & med.\ $\Delta$ IBS & IBS worse & med.\ $\Delta$ C-idx & C-idx better \\
        \midrule
        MLP & 22 & $+0.0069$ & 17/22 & $+0.0018$ & 13/22 \\
        RealMLP & 19 & $+0.0150$ & 14/19 & $+0.0018$ & 12/19 \\
        TabM & 20 & $+0.0142$ & 17/20 & $+0.0020$ & 13/20 \\
        SurvTRACE & 13 & $+0.0082$ & 9/13 & $+0.0006$ & 7/13 \\
        \midrule
        Pooled & 74 & $+0.0117$ & 57/74 (77\%) & $+0.0018$ & 45/74 (61\%) \\
        \bottomrule
    \end{tabular}
\end{table}

\subsection{CRNLL survival-head ablation}\label{sec:survivalheadablation}
The CRNLL loss uses $K+1$ heads (Sec.~\ref{sec:crlosses}), one per competing event plus a dedicated
survival head, whereas the hazard-based losses use only $K$ heads. Loss and output parameterization
are therefore two choices that vary together, and an advantage we attribute to the loss could in
principle come from the extra output instead. We separate them by training CRNLL twice on each of the
four backbones (MLP, RealMLP, TabM, SurvTRACE), once with the $(K+1)$-th head and once with only $K$
heads renormalized without a dedicated survival output, and comparing the two on IBS and on both
calibration criteria.

\begin{figure}[h!]
    \centering
    \includegraphics[width=0.9\linewidth]{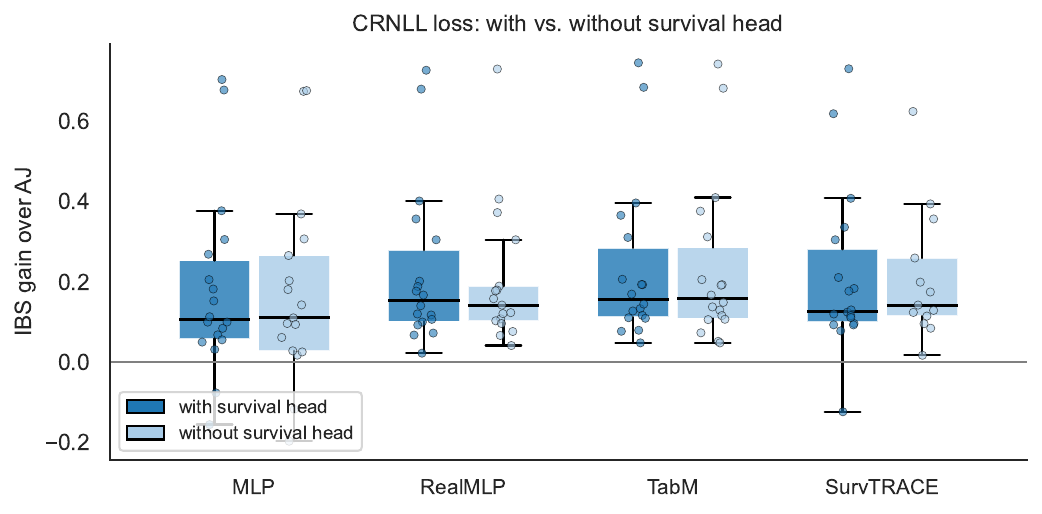}
    \caption{\textbf{The CRNLL advantage does not come from its extra survival head.} IBS gain over
    the Aalen--Johansen estimator for CRNLL trained with the dedicated $(K+1)$-th survival head (dark)
    and with $K$ heads renormalized without one (light), paired per backbone; higher is better. Each
    point is one dataset, the box spans the interquartile range and the whiskers reach 1.5 IQR. The
    two variants overlap on all four backbones and neither is systematically ahead, so ``loss'' and
    ``output parameterization'' are separable here and it is the loss that carries the effect.}
    \label{fig:survheadibs}
\end{figure}

\begin{figure}[h!]
    \centering
    \begin{subfigure}{\linewidth}
        \centering
        \includegraphics[width=0.9\linewidth]{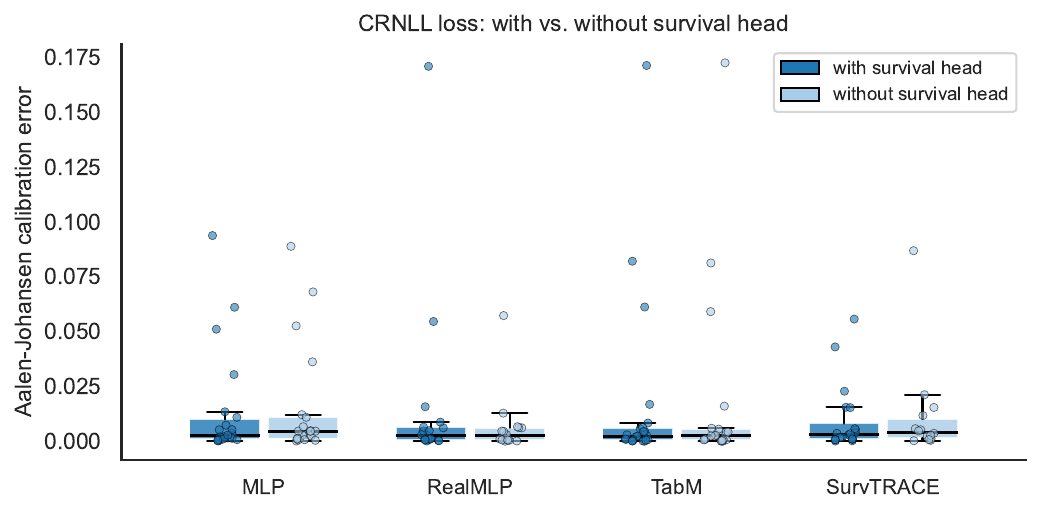}
        \caption{AJ-calibration error (lower is better).}
        \label{fig:survheadaj}
    \end{subfigure}
    \begin{subfigure}{\linewidth}
        \centering
        \includegraphics[width=0.9\linewidth]{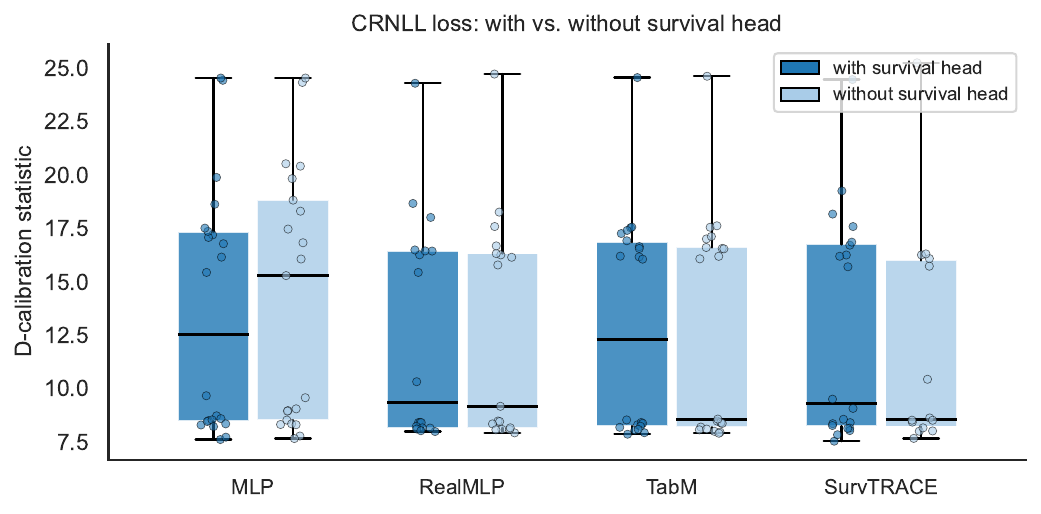}
        \caption{D-calibration statistic (lower is better).}
        \label{fig:survheadd}
    \end{subfigure}
    \caption{\textbf{The survival head does not change calibration either.} Same ablation and same
    layout as Fig.~\ref{fig:survheadibs}, scored on the two competing-risks calibration criteria of
    Sec.~\ref{sec:crmetrics}. On AJ-calibration both variants concentrate near zero on every backbone,
    with the same handful of poorly calibrated datasets in each. D-calibration is far more dispersed,
    but the paired boxes still overlap heavily and the better median alternates between the two
    variants across backbones, which is what one expects when the parameterization is immaterial
    rather than beneficial.}
    \label{fig:survheadcal}
\end{figure}

The extra head is not what makes CRNLL competitive. On IBS (Fig.~\ref{fig:survheadibs}) the two
variants overlap on every backbone: their medians differ by far less than the spread across datasets,
and the sign of the difference is not even consistent, RealMLP slightly favouring the extra head and
SurvTRACE slightly favouring its removal. Both calibration criteria behave the same way
(Fig.~\ref{fig:survheadcal}), with the paired distributions essentially indistinguishable per
backbone. Since CRNLL keeps its gain once the extra output is removed, the advantage reported in
Sec.~\ref{sec:results} follows from the loss rather than from the output parameterization.

\section{Background theory: Link between the hazard function and the CIF function}\label{sec:computations}

Here, we do the derivation to explain the link of the hazard function $\lambda_k$ of the event of interest $k$, the survival function $S$ and the CIF of the event $k$ $F_k$, and the density of the CIF ($f_k$).
\begin{align}
    & \mathbb{P}(T^\star\in [t, t + dt], \Delta^\star = k | T^\star >t, \mathbf{X} = \mathbf{x}) \, dt = \frac{\mathbb{P}(T^\star\in [t, t + dt], \Delta^\star = k | \mathbf{X} = \mathbf{x}) \, dt }{\mathbb{P}(T^\star > t | \mathbf{X} = \mathbf{x})} \\
    & \Longrightarrow \lambda_k(t|\mathbf{x}) = \frac{f_k(t|\mathbf{x})}{S(t|\mathbf{x})} \\
    & \Longrightarrow F_k(t|\mathbf{x}) = \int_0^t \lambda_k(t|\mathbf{x})S(t|\mathbf{x}) \, du
\end{align}

\end{document}